\pdfoutput=1
\documentclass[runningheads]{llncs}
\usepackage{graphicx}
\usepackage{comment}
\usepackage{amsmath,amssymb}
\usepackage{color}

\usepackage{booktabs}
\usepackage{multirow}
\usepackage{verbatim}
\usepackage{shortbold}
\usepackage{xcolor,colortbl}

\usepackage[font=small,labelfont=bf]{caption}
\usepackage{multicol}

\definecolor{DS0b0}{rgb}{0.282884,0.135920,0.453427}
\definecolor{DS0f0}{rgb}{0.900000,0.900000,0.900000}
\definecolor{DS0b1}{rgb}{0.265145,0.232956,0.516599}
\definecolor{DS0f1}{rgb}{0.900000,0.900000,0.900000}
\definecolor{DS0b2}{rgb}{0.150476,0.504369,0.557430}
\definecolor{DS0f2}{rgb}{0.900000,0.900000,0.900000}
\definecolor{DS0b3}{rgb}{0.945636,0.899815,0.112838}
\definecolor{DS0f3}{rgb}{0.000000,0.000000,0.000000}
\definecolor{DS0b4}{rgb}{0.246070,0.738910,0.452024}
\definecolor{DS0f4}{rgb}{0.900000,0.900000,0.900000}
\definecolor{DS0b5}{rgb}{0.327796,0.773980,0.406640}
\definecolor{DS0f5}{rgb}{0.000000,0.000000,0.000000}
\definecolor{DS0b6}{rgb}{0.252899,0.742211,0.448284}
\definecolor{DS0f6}{rgb}{0.900000,0.900000,0.900000}
\definecolor{DS0b7}{rgb}{1,1,1}
\definecolor{DS0f7}{rgb}{0.900000,0.900000,0.900000}
\definecolor{DS0b8}{rgb}{0.277941,0.056324,0.381191}
\definecolor{DS0f8}{rgb}{0.900000,0.900000,0.900000}
\definecolor{DS0b9}{rgb}{0.180629,0.429975,0.557282}
\definecolor{DS0f9}{rgb}{0.900000,0.900000,0.900000}
\definecolor{DS0b10}{rgb}{0.151918,0.500685,0.557587}
\definecolor{DS0f10}{rgb}{0.900000,0.900000,0.900000}
\definecolor{DS0b11}{rgb}{0.227802,0.326594,0.546532}
\definecolor{DS0f11}{rgb}{0.900000,0.900000,0.900000}
\definecolor{DS1b0}{rgb}{0.275191,0.194905,0.496005}
\definecolor{DS1f0}{rgb}{0.900000,0.900000,0.900000}
\definecolor{DS1b1}{rgb}{0.220057,0.343307,0.549413}
\definecolor{DS1f1}{rgb}{0.900000,0.900000,0.900000}
\definecolor{DS1b2}{rgb}{0.140536,0.530132,0.555659}
\definecolor{DS1f2}{rgb}{0.900000,0.900000,0.900000}
\definecolor{DS1b3}{rgb}{0.935904,0.898570,0.108131}
\definecolor{DS1f3}{rgb}{0.000000,0.000000,0.000000}
\definecolor{DS1b4}{rgb}{0.344074,0.780029,0.397381}
\definecolor{DS1f4}{rgb}{0.000000,0.000000,0.000000}
\definecolor{DS1b5}{rgb}{0.555484,0.840254,0.269281}
\definecolor{DS1f5}{rgb}{0.000000,0.000000,0.000000}
\definecolor{DS1b6}{rgb}{0.311925,0.767822,0.415586}
\definecolor{DS1f6}{rgb}{0.900000,0.900000,0.900000}
\definecolor{DS1b7}{rgb}{1,1,1}
\definecolor{DS1f7}{rgb}{0.900000,0.900000,0.900000}
\definecolor{DS1b8}{rgb}{0.280894,0.078907,0.402329}
\definecolor{DS1f8}{rgb}{0.900000,0.900000,0.900000}
\definecolor{DS1b9}{rgb}{0.150476,0.504369,0.557430}
\definecolor{DS1f9}{rgb}{0.900000,0.900000,0.900000}
\definecolor{DS1b10}{rgb}{0.194100,0.399323,0.555565}
\definecolor{DS1f10}{rgb}{0.900000,0.900000,0.900000}
\definecolor{DS1b11}{rgb}{0.258965,0.251537,0.524736}
\definecolor{DS1f11}{rgb}{0.900000,0.900000,0.900000}
\definecolor{DS2b0}{rgb}{0.258965,0.251537,0.524736}
\definecolor{DS2f0}{rgb}{0.900000,0.900000,0.900000}
\definecolor{DS2b1}{rgb}{0.190631,0.407061,0.556089}
\definecolor{DS2f1}{rgb}{0.900000,0.900000,0.900000}
\definecolor{DS2b2}{rgb}{0.147607,0.511733,0.557049}
\definecolor{DS2f2}{rgb}{0.900000,0.900000,0.900000}
\definecolor{DS2b3}{rgb}{0.993248,0.906157,0.143936}
\definecolor{DS2f3}{rgb}{0.000000,0.000000,0.000000}
\definecolor{DS2b4}{rgb}{0.281477,0.755203,0.432552}
\definecolor{DS2f4}{rgb}{0.900000,0.900000,0.900000}
\definecolor{DS2b5}{rgb}{0.140210,0.665859,0.513427}
\definecolor{DS2f5}{rgb}{0.900000,0.900000,0.900000}
\definecolor{DS2b6}{rgb}{0.128087,0.647749,0.523491}
\definecolor{DS2f6}{rgb}{0.900000,0.900000,0.900000}
\definecolor{DS2b7}{rgb}{1,1,1}
\definecolor{DS2f7}{rgb}{0.900000,0.900000,0.900000}
\definecolor{DS2b8}{rgb}{0.282656,0.100196,0.422160}
\definecolor{DS2f8}{rgb}{0.900000,0.900000,0.900000}
\definecolor{DS2b9}{rgb}{0.258965,0.251537,0.524736}
\definecolor{DS2f9}{rgb}{0.900000,0.900000,0.900000}
\definecolor{DS2b10}{rgb}{0.150476,0.504369,0.557430}
\definecolor{DS2f10}{rgb}{0.900000,0.900000,0.900000}
\definecolor{DS2b11}{rgb}{0.233603,0.313828,0.543914}
\definecolor{DS2f11}{rgb}{0.900000,0.900000,0.900000}
\definecolor{DS3b0}{rgb}{0.267004,0.004874,0.329415}
\definecolor{DS3f0}{rgb}{0.900000,0.900000,0.900000}
\definecolor{DS3b1}{rgb}{0.271305,0.019942,0.347269}
\definecolor{DS3f1}{rgb}{0.900000,0.900000,0.900000}
\definecolor{DS3b2}{rgb}{0.160665,0.478540,0.558115}
\definecolor{DS3f2}{rgb}{0.900000,0.900000,0.900000}
\definecolor{DS3b3}{rgb}{0.916242,0.896091,0.100717}
\definecolor{DS3f3}{rgb}{0.000000,0.000000,0.000000}
\definecolor{DS3b4}{rgb}{0.150148,0.676631,0.506589}
\definecolor{DS3f4}{rgb}{0.900000,0.900000,0.900000}
\definecolor{DS3b5}{rgb}{0.319809,0.770914,0.411152}
\definecolor{DS3f5}{rgb}{0.000000,0.000000,0.000000}
\definecolor{DS3b6}{rgb}{0.296479,0.761561,0.424223}
\definecolor{DS3f6}{rgb}{0.900000,0.900000,0.900000}
\definecolor{DS3b7}{rgb}{1,1,1}
\definecolor{DS3f7}{rgb}{0.900000,0.900000,0.900000}
\definecolor{DS3b8}{rgb}{0.267004,0.004874,0.329415}
\definecolor{DS3f8}{rgb}{0.900000,0.900000,0.900000}
\definecolor{DS3b9}{rgb}{0.171176,0.452530,0.557965}
\definecolor{DS3f9}{rgb}{0.900000,0.900000,0.900000}
\definecolor{DS3b10}{rgb}{0.121148,0.592739,0.544641}
\definecolor{DS3f10}{rgb}{0.900000,0.900000,0.900000}
\definecolor{DS3b11}{rgb}{0.194100,0.399323,0.555565}
\definecolor{DS3f11}{rgb}{0.900000,0.900000,0.900000}
\definecolor{DS4b0}{rgb}{0.280267,0.073417,0.397163}
\definecolor{DS4f0}{rgb}{0.900000,0.900000,0.900000}
\definecolor{DS4b1}{rgb}{0.283197,0.115680,0.436115}
\definecolor{DS4f1}{rgb}{0.900000,0.900000,0.900000}
\definecolor{DS4b2}{rgb}{0.156270,0.489624,0.557936}
\definecolor{DS4f2}{rgb}{0.900000,0.900000,0.900000}
\definecolor{DS4b3}{rgb}{0.935904,0.898570,0.108131}
\definecolor{DS4f3}{rgb}{0.000000,0.000000,0.000000}
\definecolor{DS4b4}{rgb}{0.252899,0.742211,0.448284}
\definecolor{DS4f4}{rgb}{0.900000,0.900000,0.900000}
\definecolor{DS4b5}{rgb}{0.404001,0.800275,0.362552}
\definecolor{DS4f5}{rgb}{0.000000,0.000000,0.000000}
\definecolor{DS4b6}{rgb}{0.360741,0.785964,0.387814}
\definecolor{DS4f6}{rgb}{0.000000,0.000000,0.000000}
\definecolor{DS4b7}{rgb}{1,1,1}
\definecolor{DS4f7}{rgb}{0.900000,0.900000,0.900000}
\definecolor{DS4b8}{rgb}{0.274952,0.037752,0.364543}
\definecolor{DS4f8}{rgb}{0.900000,0.900000,0.900000}
\definecolor{DS4b9}{rgb}{0.156270,0.489624,0.557936}
\definecolor{DS4f9}{rgb}{0.900000,0.900000,0.900000}
\definecolor{DS4b10}{rgb}{0.150476,0.504369,0.557430}
\definecolor{DS4f10}{rgb}{0.900000,0.900000,0.900000}
\definecolor{DS4b11}{rgb}{0.225863,0.330805,0.547314}
\definecolor{DS4f11}{rgb}{0.900000,0.900000,0.900000}

\begin{document}
\pagestyle{headings}
\mainmatter
\def\ECCVSubNumber{2820} 

\title{Full-Body Awareness from Partial Observations}

\titlerunning{Full-Body Awareness from Partial Observations}
\author{Chris Rockwell\orcidID{0000-0003-3510-5382} \and
	David F. Fouhey\orcidID{0000-0001-5028-5161}}
\authorrunning{C. Rockwell and D. F. Fouhey}
\institute{University of Michigan, Ann Arbor  \\
\email{\{cnris,fouhey\}@umich.edu}}

\maketitle

\begin{abstract}
There has been great progress in human 3D mesh recovery
and great interest in learning about the world from consumer video 
data. Unfortunately current methods for
3D human mesh recovery work rather poorly on consumer video data,
since on the Internet, unusual camera viewpoints and aggressive truncations are
the norm rather than a rarity. We study this problem
and make a number of contributions to address it: 
(i) we propose a simple but highly effective self-training framework
that adapts human 3D mesh recovery systems to consumer videos and demonstrate
its application to two recent systems;
(ii) we introduce evaluation protocols and keypoint annotations for 13K frames across four consumer
video datasets for studying this task, including evaluations on out-of-image keypoints; and 
(iii) we show that our method substantially improves 
PCK and human-subject judgments compared to baselines, both on test videos from the dataset it
was trained on, as well as on three other datasets without further adaptation.
\keywords{Human Pose Estimation}
\end{abstract}

\section{Introduction}

\begin{figure}[t]
	\centering
			{\includegraphics[width=\linewidth]{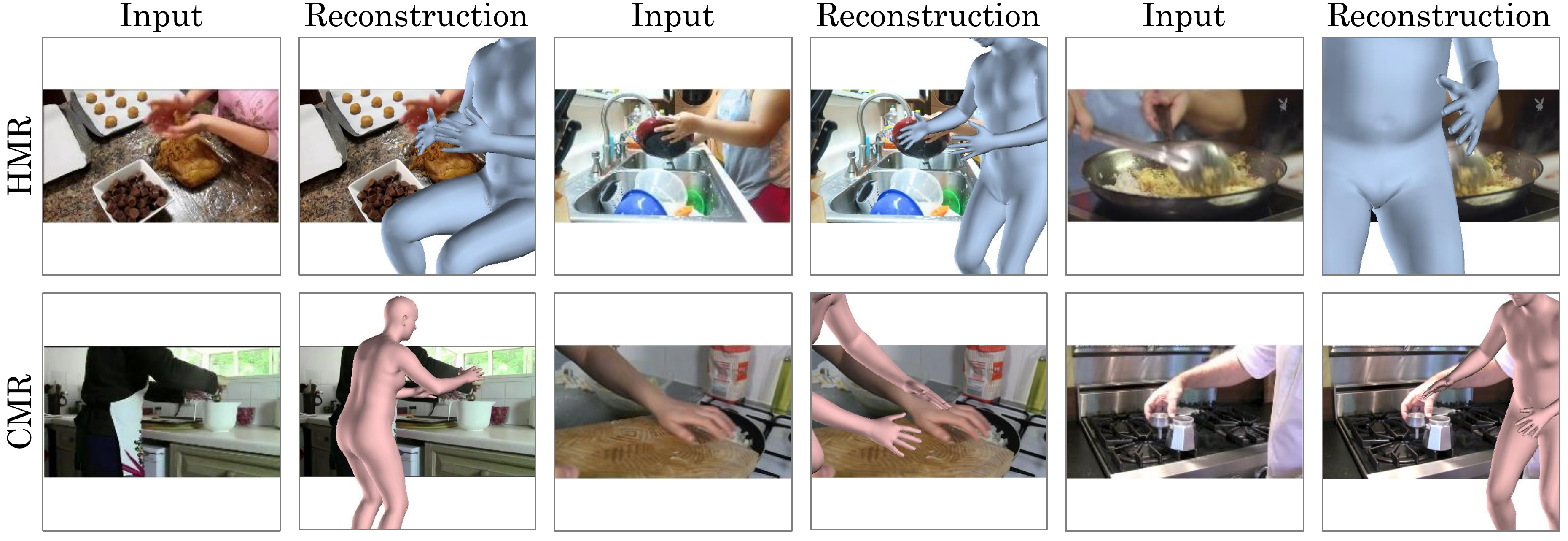}}
	\caption{We present a simple but highly effective framework for adapting human pose estimation methods to highly truncated settings
    that requires no additional pose annotation.
	We evaluate the approach on HMR \cite{KanazawaHMR17} and CMR \cite{Kolotouros19} by
 annotating four Internet video
test sets: VLOG \cite{Fouhey18} (top-left, top-middle), 
Cross-Task \cite{Zhukov19} (top-right, bottom-left), YouCookII \cite{ZhXuCoCVPR18} (bottom-middle), and Instructions \cite{Alayrac2016} (bottom-right).}

	\label{fig:fig1}
\end{figure}

Consider the images in Fig. \ref{fig:fig1}: what are these people doing? Are they standing or sitting?
While a human can readily recognize what is going on in the images, having a similar understanding
is a severe challenge to current human 3D pose estimation systems.
Unfortunately, in the world of Internet video, frames like these are the rule
rather than rarities since consumer videos are recorded not with the
goal of providing clean demonstrations of people performing poses, but are
instead meant to show something interesting to people who already know how
to parse 3D poses. 
Accordingly, while videos from consumer sharing sites
may be a useful source of data for learning how the world
works \cite{Alayrac2016,Fouhey18,ZhXuCoCVPR18,Zhukov19},
most consumer videos depict a confusing jumble of limbs and torsos
flashing across the screen. The goal of this paper is to make sense
of this jumble.

\begin{figure*}[t!]
	\centering
			{\includegraphics[width=\textwidth]{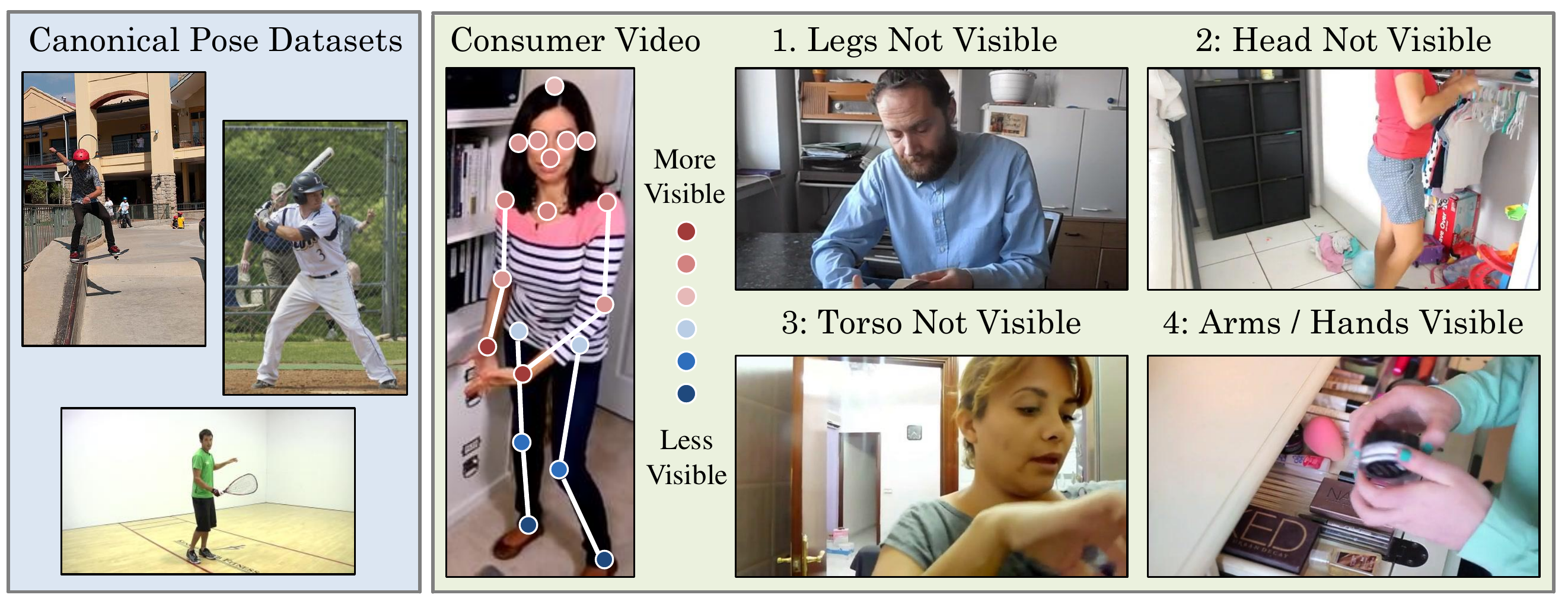}}
    \captionof{figure}{{\bf Partially Visible Humans.} Consumer video, seen in datasets like VLOG \cite{Fouhey18}, Instructions \cite{Zhukov19}, or YouCook2 \cite{ZhXuCoCVPR18}, is considerably different from canonical human pose datasets.
Most critically, only part of a person is typically visible within an image,
making pose estimation challenging.
In fact, all keypoints are only visible in 4\% of VLOG test set images, 
 while all leg joints are not visible 61\% of the time. Four of the most common configurations of visible body parts are listed above. }
	\label{fig:fig2}
\end{figure*}

Current work in human pose estimation is usually not up to the challenge of the jumble of Internet footage.
Recent work in human pose estimation \cite{alp2018densepose,Cao2017,KanazawaHMR17,martinez2017simple,newell2016stacked} is typically trained
and evaluated on 2D and 3D pose datasets
\cite{Andriluka14,Ionescu13,Johnson10,Lin2014,Mehta2017} that show full
human poses from level cameras often in athletic settings Fig.~\ref{fig:fig2} (left). Unfortunately, Internet
footage tends to be like Fig. \ref{fig:fig2} (right), and frequently only part of the body is visible to best show off
how to perform a task or highlight something of interest. For instance, on VLOG
\cite{Fouhey18}, all human joints are visible in only 4\% of image frames. Meanwhile,
all leg keypoints are {\it not} visible 63\% of the time, and head keypoints such as eyes
are {\it not} visible in about 45\% of frames. Accordingly, when standard approaches are tested
on this sort of data, they tend to fail catastrophically, which we show empirically.

We propose a simple but surprisingly effective approach in Section
\ref{sec:approach} that we apply to multiple forms of human
mesh recovery. The key insight is to combine both cropping {\it and} self-training on
confident video frames: cropping introduces the model to truncation, 
video matches context to truncations.
After pre-training on a cropped version of a
standard dataset, we identify reliable predictions on a large unlabeled video
dataset, and promote these instances to the training set and repeat. Unlike standard
self-training, we add crops, which lets confident full-body predictions
(identified via \cite{bahat2018confidence}) provide a training signal for
challenging crops. This approach requires no extra annotations and takes
$<30$k iterations of additional training (with total time $<8$
hours on a single RTX2080 Ti GPU). 

We demonstrate the effectiveness of our approach on two human 3D mesh 
recovery techniques -- HMR \cite{KanazawaHMR17} and CMR \cite{Kolotouros19} --
and evaluate on four consumer-video datasets 
-- VLOG \cite{Fouhey18}, Instructions \cite{Alayrac2016},
YouCookII \cite{ZhXuCoCVPR18}, and Cross-Task \cite{Zhukov19}.
To lay the groundwork for future work, we annotate
keypoints on 13k frames across these datasets and provide a framework for evaluation
in and out of images. 
In addition to keypoints, we evaluate using
human-study experiments. Our experiments in Section \ref{sec:experiments} demonstrate
the effectiveness of our method compared to off-the-shelf mesh recovery
and training on crops from a standard image dataset (MPII). 
Our approach improves PCK both {\it in-image} and {\it out-of-image} 
across methods and datasets: e.g., after training on VLOG, our approach
leads to a 20.7\% improvement on YouCookII over off-the-shelf HMR and a 10.9\%
improvement over HMR trained on crops (with gains of 36.4\% and 19.1\% on out-of-image keypoints)
Perceptual judgments by annotators show similar gains: e.g.,
on Cross-Task, our proposed method improves the chance
of a CMR output being rated as correct by 25.6\% compared to off-the-shelf performance.

\section{Related Work}

\par \noindent {\bf Human Pose Estimation In the Wild:} Human pose estimation has improved substantially in recent years
due in part to improved methods for 2D \cite{Cao2017,insafutdinov2016deepercut,newell2016stacked,toshev2014deeppose,xiao2018simple}
and 3D \cite{Akhter_2015_CVPR,lee1985determination,martinez2017simple,ramakrishna2012reconstructing,rogez2016mocap,zhou2016sparseness} 
pose, which typically utilize deep networks as opposed to classic approaches such as deformable part models \cite{Bourdev09,desai2012detecting,felzenszwalb2008discriminatively,Yang11}.
Performance of such pose models also relies critically on datasets \cite{Andriluka14,IonescuSminchisescu11,Ionescu13,Johnson10,Lin2014,von2018recovering,Mehta2017,Sigal2010}. 
By utilizing annotated people
\textit{in-the-wild}, methods have moved toward understanding realistic, challenging settings, and become more robust
to occlusion, setting, challenging pose, and scale variation \cite{alp2018densepose,mehta2017monocular,Mehta2017,papandreou2017towards,pavlakos2017coarse,zhou2017towards}.
 
However, these \textit{in-the-wild} datasets still rarely encounter close, varied camera angles common in consumer Internet video,
which can result in people being only partially within an image. 
Furthermore, images that do contain truncated people are sometimes filtered out \cite{KanazawaHMR17}.
As a result, the state-of-the-art on common benchmarks performs poorly in consumer videos.
In this work, we utilize the unlabeled video dataset VLOG to improve in this setting.

\vspace{1em}
\par \noindent {\bf 3D Human Mesh Estimation:} A 3D mesh is a rich representation of pose, which is employed for the method presented
in this paper. Compared to keypoints, a mesh represents a clear understanding of a person's body invariant to global 
orientation and scale. A number of recent methods \cite{bogo2016keep,KanazawaHMR17,lassner2017unite,pavlakos2018learning,tung2017self,xiang2019monocular}
build this mesh by learning to predict parametric human body models such as SMPL \cite{Loper15} or the closely-related Adam \cite{joo2018total}.
To increase training breadth,
some of these methods train on 2D keypoints \cite{KanazawaHMR17,pavlakos2018learning} and utilize a shape prior.

The HMR \cite{KanazawaHMR17} model trains an adversarial prior with a variety of 2D keypoint datasets to demonstrate good performance \textit{in-the-wild}, 
making it a strong candidate to extend to more challenging viewpoints. 
CMR \cite{Kolotouros19} also produces strong results using a similar \textit{in-the-wild} training methodology.
We therefore apply our method to both models, rapidly improving performance on Internet video.

\vspace{1em}
\par \noindent {\bf Understanding Partially-Observed People:} Much of the prior work studying global understanding of partially-observed people comes from 
ego-centric action recognition \cite{fathi2011understanding,li2015delving,ma2016going,ryoo2013first,sigurdsson2016hollywood}. 
Methods often use observations of the same human body-parts between images, typically hands \cite{fathi2011understanding,li2015delving,ma2016going}, 
to classify global activity. In contrast, our goal is to predict pose, from varied viewpoints.

Some recent work explores ego-centric pose estimation. 
Recent setups use cameras mounted in a variety of clever ways, such as chest \cite{jiang2017seeing,rogez2015first}, bike helmet \cite{rhodin2016egocap}, VR goggles \cite{tome2019xr}, 
and hat \cite{xu2019mo}.
However, these methods rely on camera always being in the same spot relative to the human to make predictions.
On the other hand, our method attains global understanding of the body by training on entire people to reason
about unseen joints as it encounters less visible images.

Prior work also focuses on pose estimation specifically in cases of occlusion \cite{ghiasi2014parsing,haque2016towards}.
While this setting requires inference of non-visible joints, it does not face the same scale variation occurring in consumer video, which
can contain people much larger than the image. 
Some recent work directly addresses truncation. Vosoughi and Amer predict truncated 3D keypoints on random crops of Human3.6M \cite{saeid2018deep}.
In concurrence with our work, Exemplar Fine-Tuning \cite{joo2020exemplar} uses upper-body cropping to improve
performance in Internet video \cite{von2018recovering}.
Nevertheless, consumer Internet video (Fig.~\ref{fig:fig2}) faces more extreme truncation. 
We show cropping alone is not sufficient for this setting; rather
cropping {\it and} self-training on confident video frames provides the best results.

\section{Approach} \label{sec:approach}

\begin{figure}[t!]
	\centering
			{\includegraphics[width=\linewidth]{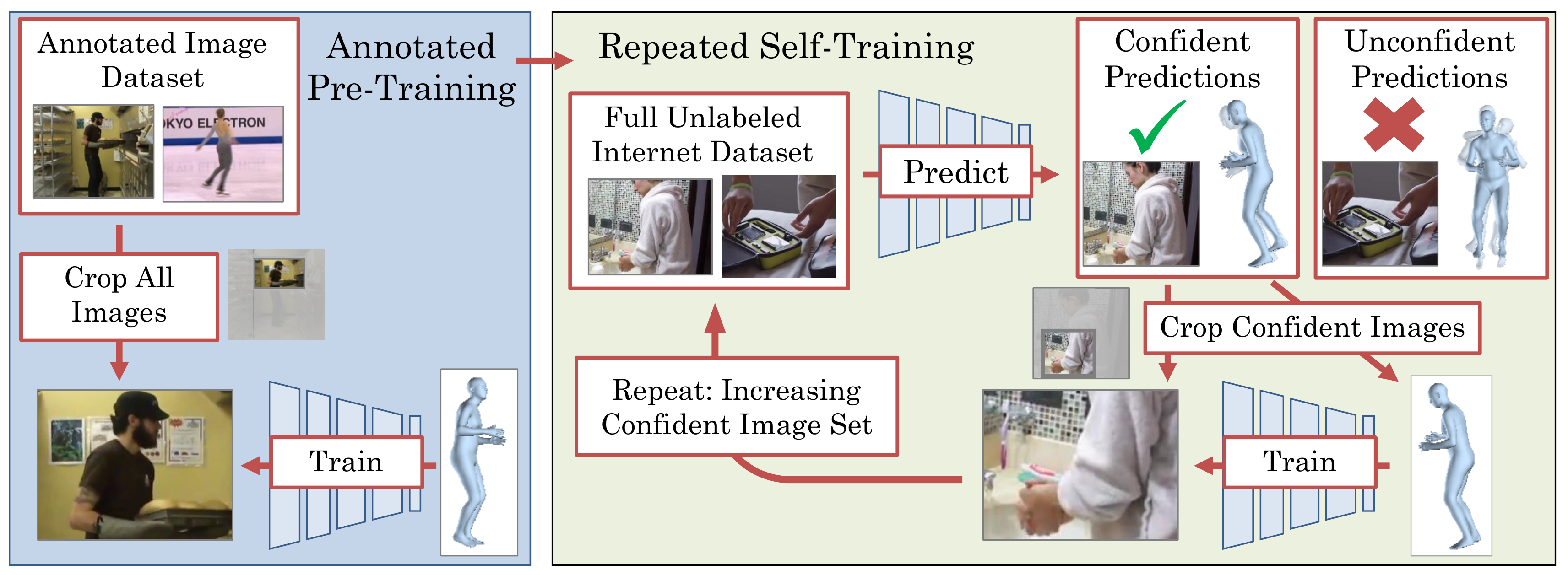}}
\caption{Our method adapts human pose models to truncated settings by self-training on cropped images.
After pre-training using an annotated pose dataset, the method applies small translations to an unlabeled 
video dataset and selects predictions with consistent pose predictions across translations as pseudo-ground-truth. 
Repeating the process increases the training set to include more truncated people.}
	\label{fig:fig4}
\end{figure}

Our goal is the ability to reconstruct a full human-body 3D mesh from an image 
of part or all of a person in consumer video data. We demonstrate how to do this
using a simple but effective self-training approach that we apply to two 3D human
mesh recovery models, HMR \cite{KanazawaHMR17} and CMR \cite{Kolotouros19}.
Both systems can predict a mesh by regressing SMPL \cite{Loper15} parameters from
which a human mesh can be generated, but work poorly on this consumer video data.

Our method, shown in Fig. \ref{fig:fig4}, adapts each method to this
challenging setting of partial visibility by sequentially self-training on
confident mesh and keypoint predictions. Starting with a model trained
on crops from a labeled dataset, the system makes predictions on 
video data. We then identify confident predictions using the equivariance technique
of Bahat and Shakhnarovich \cite{bahat2018confidence}. Finally, using the confident examples as
pseudo-ground-truth, the model is trained to map crops of the confident
images to the full-body inferences, and the process of identifying confident 
images and folding them into a training set is continued. Our only assumption
is that we can identify frames containing a single person (needed for training
HMR/CMR). In this paper, we annotate this for simplicity but assume this can be
automated via an off-the-shelf detection system.

\subsection{Base Models}

Our base models \cite{KanazawaHMR17,Kolotouros19} use SMPL \cite{Loper15}, which is a differentiable,
 generative model of human 3D meshes. SMPL maps 
parameters $\Theta$ to output a triangulated mesh $\YB$ with ${N = 6980}$ 
vertices. $\Theta$ consists of parameters 
$[{\thetaB,\betaB,\RB,\tB,\sB}]$: joint rotations $\thetaB \in \mathbb{R}^{69}$, shape
parameters $\betaB \in \mathbb{R}^{10}$, global rotation $\RB$, global translation $\tB$, and global scale $\sB$. 
We abstract each base model as a function $f$ mapping an image $I$ to a SMPL parameter $\Theta$.
As described in \cite{KanazawaHMR17}, the SMPL parameters can be used to yield a set
of 2D projected keypoints $\xB$.

Our training process closely builds off the original methods:
we minimize a sum of losses on a combination of projected 2D keypoints $\hat{\xB}$, predicted vertices $\hat{\YB}$, and SMPL parameters ${\hat{\Theta}}$.
The most important distinctions
are, we assume we have access to 
SMPL parameters  $\Theta$ for each image, and we train on all annotated
keypoints, even if they are outside the image. 
We describe salient differences between the models and original training below.

\noindent {\bf HMR \cite{KanazawaHMR17}:}
Kanazawa \textit {et al.} use MoSh \cite{loper2014mosh,varol2017learning} for their ground truth SMPL loss.
However, this data is not available in most images, and thus the model relies primarily on keypoint loss.
Instead, we train directly on predicted SMPL rotations, available in all images; we find $L_1$ loss works best.
To encourage our network to adapt to poses of new datasets, we do not use a discriminator loss. 
We also supervise ($\tB, \sB$), though experiments indicated this did not impact performance (less than 1\% difference on keypoint results).
The loss for a single datapoint is:
\begin{equation}
    \label{eqn:objective}
	L = \left\|[\thetaB, \RB, \betaB, \tB,\sB] - [\hat{\thetaB}, \hat{\RB}, \hat{\betaB}, \hat{\tB},\hat{\sB}]\right\|_1 + \left\|\xB - \hat{\xB}\right\|_1
\end{equation}

\noindent {\bf CMR \cite{Kolotouros19}:} CMR additionally regresses predicted mesh, and 
has intermediate losses after the Graph CNN.
We do not change their loss, other than by always using 3D supervision and out-of-image keypoints. It is distinct from our HMR loss as it uses $L_2$ loss on keypoints and SMPL parameters, and converts $\thetaB$ and $\RB$ to rotation matrices for training \cite{omran2018neural}, 
although they note conversion does not change quantitative results.
The loss for a single datapoint is:
\begin{equation}
    \label{eqn:objective2}
L = \left\|[\thetaB, \RB] - [\hat{\thetaB}, \hat{\RB}]\right\|_2^2 + \lambda \left\|\betaB - \hat{\betaB}\right\|_2^2 + \left\|\xB - \hat{\xB}\right\|_2^2 + \left\|\YB - \hat{\YB}\right\|_1
\end{equation}
such that $\lambda = 0.1$, each norm is reduced by its number of elements, and keypoint and mesh losses are also applied after the Graph CNN. While Kolotouros \textit{et al.}
train the Graph CNN before the MLP, we find the pretrained model trains well with both losses simultaneously.

\subsection{Iterative Adaptation to Partial Visibility}

Our approach follows a standard self-training approach to semi-supervised
learning. In self-training, one begins with an {\bf initial model} 
$f_0: \mathcal{X} \to \mathcal{Y}$ as well as a collection of unlabeled
data $U = \{u:u \in \mathcal{X}\}$. Here, the inputs are images, outputs
SMPL parameters, and model either CMR or HMR. The key idea is to use the inferences of each
round's model $f_i$ to produce labeled data for training the next round's 
model $f_{i+1}$. More specifically, at each iteration $t$, the model $f_t$
is applied to each element of $U$, and a {\bf confident prediction subset} $C \subseteq U$ is identified.
Then, predictions of model $f$ on elements are treated as new ground-truth for training
the next round model $f_{i+1}$.
In standard self-training, the new training set is
the original unlabeled inputs and model outputs, or $\{(c,f_i(c)): c \in C\}$. In our case, this would
never learn to handle more cropped people,
and the training
set is thus augmented with {\bf transformations} of the confident samples, or 
$\{(t(c),t(f_i(c))): c \in C, t \in T\}$ for some set of crops $T$. 
The new model $f_{i+1}$ is {\bf retrained} and the process is repeated until convergence. 
We now describe more concretely what
we mean by each bolded point.

\noindent {\bf Initial Model:} 
We begin by training the pretrained HMR and CMR models on MPII (Fig. \ref{fig:fig4}, left) such that 
we apply cropping transformations to images and keypoints. 
SMPL predictions from full images are used for supervision, and are typically very accurate considering past training on this set.
This training scheme is the same as that used for self-training (Fig. \ref{fig:fig4}, right), except we use MPII
ground truth keypoints instead of pseudo-ground truths.

\noindent {\bf Identifying Confident Predictions:}
In order to apply self-training, we need to be able to 
find confident outputs of each of our SMPL-regressing models.
Unfortunately, it is difficult to extract confidence from regression models
because there is no natural and automatically-produced confidence measure 
unlike in classification where measures like entropy provide a
starting point.

We therefore turn to an empirical result of Bahat and Shakhnarovich
\cite{bahat2018confidence} that invariance to image
transformations is often indicative of confidence in neural networks. Put simply,
confident predictions of networks tend to be more invariant to small transformations
(e.g., a shift) than non-confident predictions. We apply this technique
in our setting by examining changes of parameters after applying small
translational jitter: we apply the model $f$ to copies of the image with
the center jittered 10 and 20 pixels and look at joint rotation parameters 
$\thetaB$. We compute the variance of each joint rotation parameter across
the jittered samples, then average the variances across joints. For HMR, we
define confident samples as ones with a variance below $0.005$ (chosen
empirically).
For CMR,
for simplicity, we ensure that we have the same acceptance rate
as HMR of 12\%; this results in a similar variance threshold of $0.004$.

\noindent {\bf Applying Transformations:} The set of inputs and confident pseudo-label outputs that can be used for
self-training is not 
enough.
We therefore
apply a family of crops that mimic empirical frequencies found in consumer video data.
Specifically, crops consist of 23\% most of body visible, 
29\% legs not visible, 10\% head not visible, and 22\% only hands or arms visible.
Examples of these categories are shown in Fig. \ref{fig:fig2}. Although proportions
were chosen empirically from VLOG, other consumer Internet video datasets considered 
\cite{Alayrac2016,ZhXuCoCVPR18,Zhukov19} exhibit similar visibility patterns, 
and we empirically show that our results generalize.

\noindent {\bf Retraining:} Finally, given the set of samples of crops of confident images
and corresponding full bodies, we retrain each model.

\subsection{Implementation Details}
\label{subsec:impl}

Our cropping procedure and architecture is detailed in supplemental for both HMR
\cite{KanazawaHMR17} and CMR \cite{Kolotouros19}; we initialize both with weights pretrained on \textit{in-the-wild} data.
On MPII, we continue using the same learning rate and optimizer
used by each model (1e-5 for HMR, 3e-4 for CMR, both use Adam \cite{kingma2014adam}) until validation loss converges.
Training converges within 20k iterations in both cases.

Next, we identify confident predictions as detailed above on VLOG. We use the subset of the hand-contact state dataset containing single humans, which consists of 132k frames in the train + validation set.
We note we could have used a simple classifier to filter by visible people in a totally unlabeled setting.
Our resulting confident train + validation set is 15k images. 
We perform the same cropping transformations as in MPII, and continue training with the same parameters. Validation
loss converges within 10k iterations. We repeat this semi-supervised component one additional time, and the new train + validation
set is of approximately size 40k. Training again takes less than 10k iterations.

\definecolor{mygreen}{RGB}{84, 130, 53}
\definecolor{myred}{RGB}{255, 0, 0}
\begin{figure*}[t!]
	\centering
			{\includegraphics[width=\textwidth]{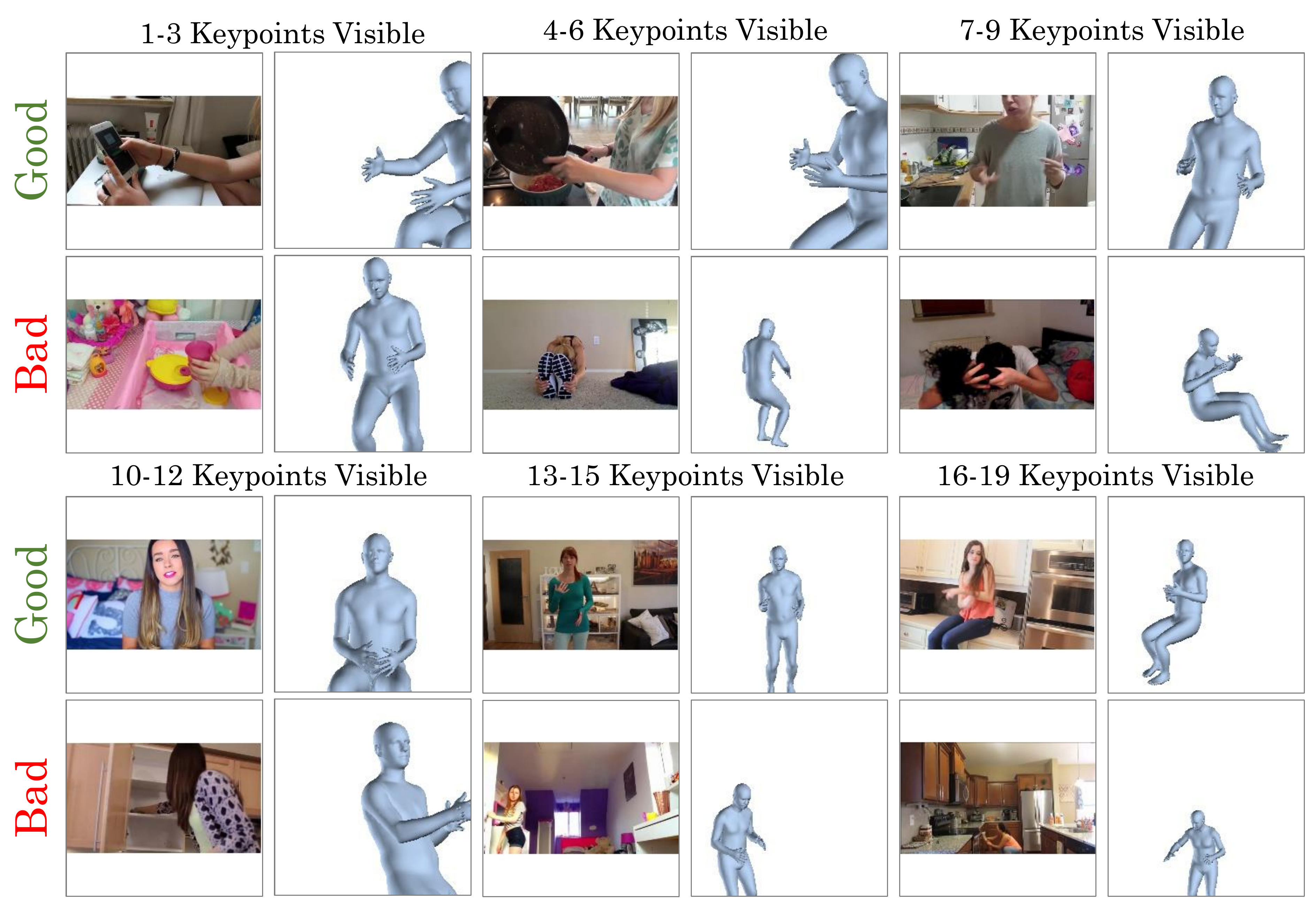}}
    \captionof{figure}{Randomly sampled \textbf{\textcolor{mygreen}{positive}} and \textbf{\textcolor{myred}{negative}} predictions by number of keypoints visible, 
    as identified by workers.
	Images with fewer keypoints visible are typically more difficult, and our method 
	improves most significantly in these cases (Table \ref{tab:good_vlog}).}
\label{fig:fig7}
\end{figure*}

\section{Experiments} \label{sec:experiments}

We now describe a set of experiments done to investigate the following experimental questions:
(1) how well do current 3D human mesh recovery systems work on consumer internet video?
(2) can we improve the performance of these systems, both on an absolute basis and in comparison
to alternate simple models that do not self-train?
We answer these questions by evaluating the performance of a variety of adaptation methods
applied to both HMR \cite{KanazawaHMR17} and CMR \cite{Kolotouros19} on four independent datasets of Internet videos.
After introducing the datasets (Sec. \ref{subsec:dataset}) and experimental
setup (Sec.~\ref{subsec:setup}), we describe experiments on VLOG (Sec.~\ref{subsec:vlog}), which we use for
our self-training video adaptation. To test the generality of our conclusions,
we then repeat the same experiments on three other consumer datasets without any retraining
(Sec.~\ref{subsec:generalization}). We validate our choice of confidence against two other methods
in (Sec.~\ref{subsec:confidence}).

\subsection{Datasets and Annotations}
\label{subsec:dataset}

\begin{table}[t]
    \caption{Joint visibility statistics across the four consumer video datasets that we use.
    Across multiple consumer video datasets, fully visible people are exceptionally rare ($<3\%$),
    in contrast to configurations like an upper torso or only pieces of someone's arms, or 
    much of a body but no head. Surprisingly, the most likely to be visible joint is actually
    wrists, more than $2$x more likely than hips and $5$x more likely than knees.}
    \label{tab:jointstatistics}
    \resizebox{\columnwidth}{!}{\begin{tabular}{cccccccccccccccccccc} \toprule & \multicolumn{7}{c}{Independent Joint Statistics} & & \multicolumn{4}{c}{Joint Joint Statistics}\\ &   &   &   &   &   & Neck +  & Head +  &   & Fully  & Upper  & Only  & All But \\ 
 & Ankle  & Knee  & Hip  & Wrist  & Elbow  & Should.  & Face  &   & Visible  & Torso  & Arms  & Head \\ 
\midrule 
Average  &  \cellcolor{DS0b0} \textcolor{DS0f0}{ 7.0} &  \cellcolor{DS0b1} \textcolor{DS0f1}{ 12.9} &  \cellcolor{DS0b2} \textcolor{DS0f2}{ 31.9} &  \cellcolor{DS0b3} \textcolor{DS0f3}{ 71.9} &  \cellcolor{DS0b4} \textcolor{DS0f4}{ 50.8} &  \cellcolor{DS0b5} \textcolor{DS0f5}{ 53.9} &  \cellcolor{DS0b6} \textcolor{DS0f6}{ 51.1} &  \cellcolor{DS0b7} \textcolor{DS0f7}{ ~~ } &  \cellcolor{DS0b8} \textcolor{DS0f8}{ 2.8} &  \cellcolor{DS0b9} \textcolor{DS0f9}{ 26.2} &  \cellcolor{DS0b10} \textcolor{DS0f10}{ 31.8} &  \cellcolor{DS0b11} \textcolor{DS0f11}{ 18.8}\\ 
\midrule
VLOG \cite{Fouhey18}  &  \cellcolor{DS1b0} \textcolor{DS1f0}{ 10.5} &  \cellcolor{DS1b1} \textcolor{DS1f1}{ 20.0} &  \cellcolor{DS1b2} \textcolor{DS1f2}{ 34.0} &  \cellcolor{DS1b3} \textcolor{DS1f3}{ 71.5} &  \cellcolor{DS1b4} \textcolor{DS1f4}{ 54.5} &  \cellcolor{DS1b5} \textcolor{DS1f5}{ 61.0} &  \cellcolor{DS1b6} \textcolor{DS1f6}{ 53.3} &  \cellcolor{DS1b7} \textcolor{DS1f7}{ ~~ } &  \cellcolor{DS1b8} \textcolor{DS1f8}{ 4.0} &  \cellcolor{DS1b9} \textcolor{DS1f9}{ 32.0} &  \cellcolor{DS1b10} \textcolor{DS1f10}{ 24.0} &  \cellcolor{DS1b11} \textcolor{DS1f11}{ 14.0}\\ 
Instructions \cite{Alayrac2016}  &  \cellcolor{DS2b0} \textcolor{DS2f0}{ 14.0} &  \cellcolor{DS2b1} \textcolor{DS2f1}{ 24.5} &  \cellcolor{DS2b2} \textcolor{DS2f2}{ 32.5} &  \cellcolor{DS2b3} \textcolor{DS2f3}{ 73.5} &  \cellcolor{DS2b4} \textcolor{DS2f4}{ 52.0} &  \cellcolor{DS2b5} \textcolor{DS2f5}{ 44.7} &  \cellcolor{DS2b6} \textcolor{DS2f6}{ 43.3} &  \cellcolor{DS2b7} \textcolor{DS2f7}{ ~~ } &  \cellcolor{DS2b8} \textcolor{DS2f8}{ 5.0} &  \cellcolor{DS2b9} \textcolor{DS2f9}{ 14.0} &  \cellcolor{DS2b10} \textcolor{DS2f10}{ 32.0} &  \cellcolor{DS2b11} \textcolor{DS2f11}{ 18.0}\\ 
YouCook II \cite{ZhXuCoCVPR18}  &  \cellcolor{DS3b0} \textcolor{DS3f0}{ 0.0} &  \cellcolor{DS3b1} \textcolor{DS3f1}{ 1.0} &  \cellcolor{DS3b2} \textcolor{DS3f2}{ 30.0} &  \cellcolor{DS3b3} \textcolor{DS3f3}{ 71.0} &  \cellcolor{DS3b4} \textcolor{DS3f4}{ 45.5} &  \cellcolor{DS3b5} \textcolor{DS3f5}{ 53.7} &  \cellcolor{DS3b6} \textcolor{DS3f6}{ 52.7} &  \cellcolor{DS3b7} \textcolor{DS3f7}{ ~~ } &  \cellcolor{DS3b8} \textcolor{DS3f8}{ 0.0} &  \cellcolor{DS3b9} \textcolor{DS3f9}{ 28.0} &  \cellcolor{DS3b10} \textcolor{DS3f10}{ 39.0} &  \cellcolor{DS3b11} \textcolor{DS3f11}{ 24.0}\\ 
Cross-Task \cite{Zhukov19}  &  \cellcolor{DS4b0} \textcolor{DS4f0}{ 3.5} &  \cellcolor{DS4b1} \textcolor{DS4f1}{ 6.0} &  \cellcolor{DS4b2} \textcolor{DS4f2}{ 31.0} &  \cellcolor{DS4b3} \textcolor{DS4f3}{ 71.5} &  \cellcolor{DS4b4} \textcolor{DS4f4}{ 51.0} &  \cellcolor{DS4b5} \textcolor{DS4f5}{ 56.3} &  \cellcolor{DS4b6} \textcolor{DS4f6}{ 55.0} &  \cellcolor{DS4b7} \textcolor{DS4f7}{ ~~ } &  \cellcolor{DS4b8} \textcolor{DS4f8}{ 2.0} &  \cellcolor{DS4b9} \textcolor{DS4f9}{ 31.0} &  \cellcolor{DS4b10} \textcolor{DS4f10}{ 32.0} &  \cellcolor{DS4b11} \textcolor{DS4f11}{ 19.0}\\ 
\bottomrule \end{tabular}} 
\end{table}

We rely on four datasets of consumer video from the Internet for evaluating our method:
VLOG \cite{Fouhey18}, Instructions \cite{Alayrac2016},
YouCookII \cite{ZhXuCoCVPR18}, and Cross-Task \cite{Zhukov19}. 
Evaluation on VLOG takes place on a random 5k image subset of the test set detailed in
Sec. \ref{subsec:impl}. 
For evaluation on Instructions, YouCookII, and Cross-Task, 
we randomly sample test-set frames (Instructions we sample from the entire dataset, which is used for cross-validation),
which are filtered via crowd-workers by whether there is a single person, and then randomly subsample 5k subset.

Finally, to enable automatic metrics like PCK, we obtain joint annotations 
on all four datasets. We annotate keypoints for the 19 joints reprojected from HMR, or
the 17 COCO keypoints along with the neck and head top from MPII.
Annotations are crowd-gathered by workers who must pass a qualification test and are monitored by sentinels, and is 
detailed in supplemental.
We show statistics of these joints in Table~\ref{tab:jointstatistics}, which show quantitatively
the lack of visible keypoints. In stark contrast to canonical pose datasets, the head is often not visible.
Instead, the most frequently visible joints are wrists. 

\subsection{Experimental Setup}
\label{subsec:setup}

We evaluate our approaches as well as a set of baselines that test concrete hypotheses 
using two styles of metrics: 2D keypoint metrics, specifically PCK
measured on both in-image joints as well as via out-of-image joints (via evaluation on crops); and
3D Mesh Human Judgments, where crowd workers evaluate outputs of the systems on an absolute
or relative basis.

\noindent {\bf 2D Keypoint Metrics:} 
Our first four metrics compare predicted keypoints with annotated ones.
Our base metric is PCK {@} 0.5 \cite{Andriluka14}, the percent of
keypoints within a threshold of 0.5 times head segment,
the most commonly reported threshold on MPII. 
Our first metric, {\it Uncropped
PCK}, is performance on images where the head is visible to define PCK.
We choose PCK since head segment length
is typically undistorted in our data, as opposed to alternates where
identifying a stable threshold is difficult: PCP \cite{ferrari2008progressive} 
is affected by our high variance in
body 3D orientation, and PCPm \cite{Andriluka14} by high inter-image scale variation. 

PCK is defined only on images where the head is visible (a shortcoming
we address with human judgment experiments). In addition to being a subset, these frames 
are not representative of typical visibility patterns in consumer video 
(as shown in Fig. \ref{fig:fig2} and Table~\ref{tab:jointstatistics}), 
so we evaluate on crops.
We sample crops to closely match the joint visibility statistics of
each entire annotated test set (detailed in supplemental). We can then evaluate {\it In-Image PCK}, or PCK on joints
in the cropped image. Because the original image contains precise annotations of 
joints not visible in the crop, we can also evaluate {\it Out-of-Image PCK}, or PCK on joints outside the crop.
{\it Total PCK} is PCK on both. 
We calculate PCK on each image and then average over images. Not doing this gives
significantly more weight to images with many keypoints in them, and ignores images with few. 

\noindent {\bf 3D Mesh Human Judgments:} 
While useful, keypoint metrics like PCK suffer from a number of shortcomings.
They can only be evaluated on a subset of images: this ranges from 37\% of images from Instructions
to 50\% of images in Cross-Task. Moreover, these subsets are not necessarily representative, as argued
before. Finally, in the case of out-of-image keypoints, PCK does not distinguish between
plausible predictions that happen to be incorrect according to the fairly
exacting PCK metric, and implausible guesses. We therefore turn to human
judgments, measuring results in absolute and comparative terms.

\noindent {\it Mesh Score/Absolute Judgment:} We show workers an image and single mesh, and
ask them to classify it as largely correct or not (precise definition in
supplemental), from which we can calculate Percentage of Good Meshes: the proportion of predicted meshes workers consider good. 
Predictions from all methods are aggregated and randomly ordered, so are evaluated by the same pool of workers. 

\noindent {\it Relative Judgment:} As a verification, we also perform A/B testing on HMR
predictions. We follow a standard A/B paradigm and show
human workers an image and two meshes in random order and ask which matches the image better with
the option of a tie; when workers cannot agree, we report this as a
tie. 

\noindent {\bf Baselines:} We compare our proposed model with two baselines to answer a few scientific questions.

\noindent {\it Base Method:} We compare with
the base method being used, either HMR \cite{KanazawaHMR17} or CMR
\cite{Kolotouros19}, without any further training past their
original pose dataset training sets. This both quantifies how well
3D pose estimation methods work on consumer footage and identifies when
our approach improves over this model.

\noindent {\it Crops:} We also compare with
a model trained on MPII Crops (including losses on out-of-image keypoints). This
tests whether simply training the model on crops is sufficient 
compared to also self-training on Internet video.

\subsection{Results on VLOG}
\label{subsec:vlog}

\begin{figure*}[t!]
	\centering
			{\includegraphics[width=\textwidth]{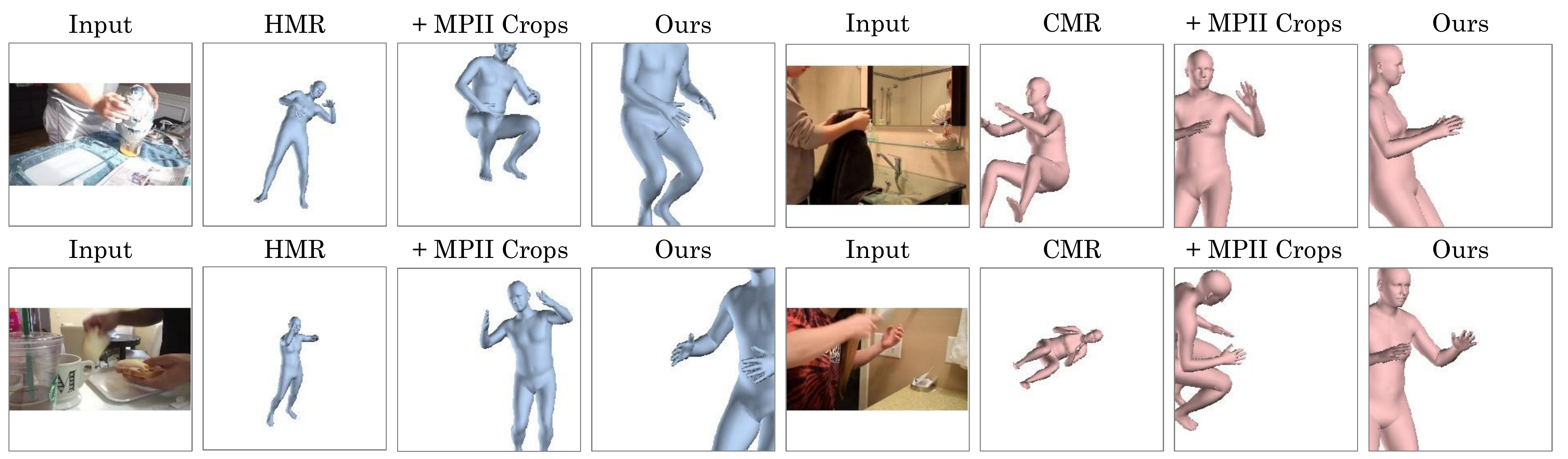}}
	\captionof{figure}{Selected comparison of results on VLOG \cite{Fouhey18}. We demonstrate sequential improvement between ablations on HMR (left) and CMR (right). Training on MPII Crops prepares the model
for truncation, while self-training provides context clues it can associate with full-body pose, leading to better predictions, particularly outside images.}

\label{fig:fig5}
\end{figure*}

\begin{figure}[t]
	\centering
			{\includegraphics[width=\linewidth]{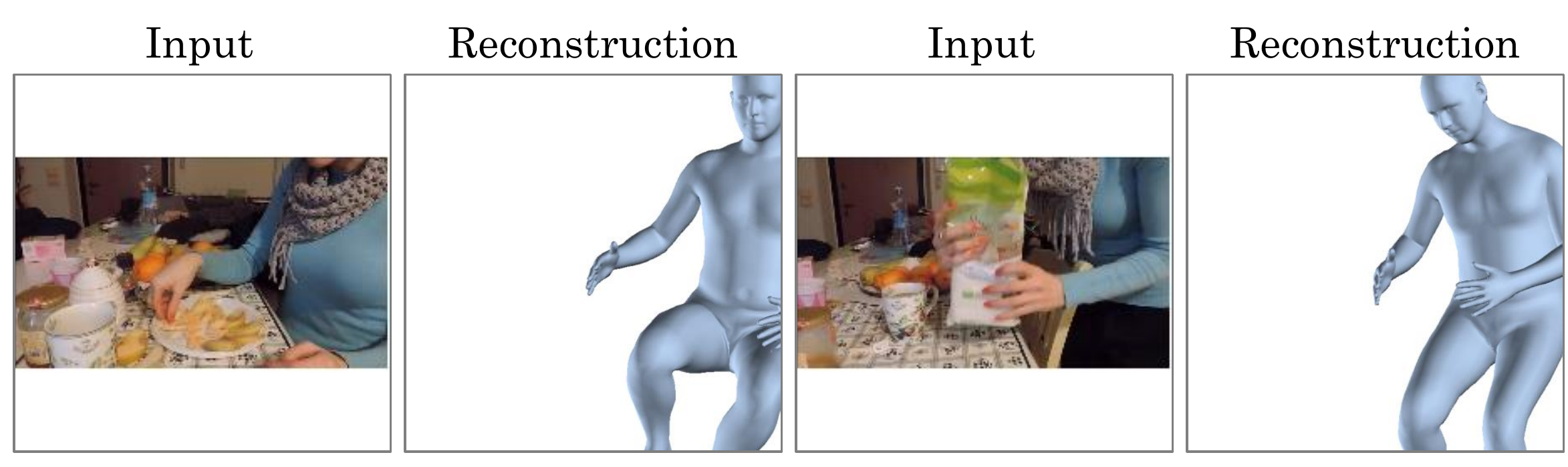}}
	\captionof{figure}{Shots focused on hands occur often in consumer video. While the visible body may look similar across instances, full-body pose can vary widely, 
	meaning keypoint detection is not sufficient for full-body reasoning. 
	After self-training,
	our method learns to differentiate activity such as standing and sitting 
	given similar visible body.
	}

\label{fig:fig11}
\end{figure}

Our first experiments are on VLOG \cite{Fouhey18}, the dataset that we train on.
We begin by showing qualitative results, comparing our method with a number
of baselines in Fig. \ref{fig:fig5}. While effective on full-body cases, 
the initial methods perform poorly on truncated people.
Training on MPII Crops prepares the model to better identify truncated people, but self-training on Internet video  
provides the model context clues it can associate with people largely outside of images ---
some of the largest improvements occur when
key indicators such as sinks and tables (Fig. \ref{fig:fig5}) are present.
In Fig. \ref{fig:fig11}, the model identifies distinct leg and head poses outside 
of images given minute difference in visible pose and appearance.

\noindent {\bf Human 3D Mesh Judgments:}
We then consider human 3D Mesh Judgments, which quantitatively confirm the
trends observed in the qualitative results. We report the frequency that each
method's predictions were rated as largely correct on the test set, broken
down by the number of visible joints, 
in Table \ref{tab:good_vlog}. Our approach {\it always}
outperforms using the base method, and only is outperformed by Crops on 
full or near-full keypoint visibility. These performance gains are particularly 
strong in the less-visible cases compared to both the base method and crops.
For instance, by using our technique, HMR's performance in highly truncated
configurations (1-3 Keypoints Visible) is improved by 23.7 points compared to the base and 11.0 
compared to using crops. 

\begin{table}[t]
    \caption{{\bf Percentage of Good Meshes on VLOG}, as judged by human workers. We report results on {\it All} images and examine results by number of visible keypoints.}
    \label{tab:good_vlog}
\resizebox{\ifdim\width>\columnwidth
        \columnwidth
      \else
        \width
      \fi}{!}{
\begin{tabular}{l @{~~~~} ccccccc @{~~~~} ccccccc}
\toprule
        & \multicolumn{7}{c}{HMR \cite{KanazawaHMR17}} & \multicolumn{7}{c}{CMR \cite{Kolotouros19}}
\\
        & \multicolumn{6}{c}{By \# of Visible Joints} & 
        & \multicolumn{6}{c}{By \# of Visible Joints} & 
\\
        & 1-3   & 4-6   & 7-9   & 10-12 & 13-15 & 16-19 & All
        & 1-3   & 4-6   & 7-9   & 10-12 & 13-15 & 16-19 & All
\\
\midrule
Base    & 19.2 & 52.7 & 70.1 & 80.1 & 85.2 & 82.1 & 60.6
        & 13.6 & 37.9 & 53.4 & 68.0 & 79.0 & 74.8 & 51.1
\\
Crops   & 31.9 & 68.7 & 76.9 & 86.8 & 91.0 & \bf 85.9 & 69.4
        & 33.7 & 65.8 & 75.7 & 82.5 & 88.4 & \bf 80.9 & 67.5
\\
Full    & \bf 42.9 & \bf 72.1 & \bf 82.8 & \bf 89.6 & \bf 92.3 & 83.1 & \bf 73.9
        & \bf 40.9 & \bf 71.2 & \bf 80.2 & \bf 86.0 & \bf 89.2 & 79.5 & \bf 71.2
\\ \bottomrule
\end{tabular} 
}
\end{table}

\begin{table}[t]
\caption{{\bf PCK @ 0.5 on VLOG}. We compute PCK on the 1.8k image VLOG test set, 
in which the head is fully visible, as {\it Uncr. Total}.
These images are then {\it Cropped} to emulate the keypoint
visibility statistics of the entire dataset, on which we can
calculate PCK {\it In} and {\it Out} of cropped images, and their union {\it Total}.
}
    \label{tab:pckvlog}
\setlength{\tabcolsep}{5pt} 
\centering 
\resizebox{\ifdim\width>\columnwidth
        \columnwidth
      \else
        \width
      \fi}{!}{
  \begin{tabular}{ l @{~~~~~~} c c c c @{~~~~~~} c c c c} \toprule
Method  & \multicolumn{4}{c}{HMR \cite{KanazawaHMR17}} 
        & \multicolumn{4}{c}{CMR \cite{Kolotouros19}}
        \\
        & \multicolumn{3}{c}{Cropped} & Uncr.     
        & \multicolumn{3}{c}{Cropped} & Uncr.     
        \\
        & Total   & In   & Out   & Total              
        & Total   & In   & Out   & Total              
        \\ \midrule
Base    & 48.6    & 65.2    & 14.7  & 68.5      
        & 36.1    & 50.2    & 13.2  & 49.5      
\\ 
Crops 
        & 51.6    & \bf 65.3    & 24.2  & \bf 68.8
        & 47.3	  & 58.1    & 26.2  & \bf 59.5
\\
Ours         
        & \bf 55.9    & 61.6    & \bf 38.9  & 68.7
        & \bf 50.9    & \bf 60.3    & \bf 34.6  & 58.1
\\ \bottomrule
\end{tabular}
\vspace{-0.1in}
}
\label{tab:pck_combined}
\end{table}

\noindent {\bf 2D Keypoints:}
We next evaluate keypoints, reporting results for all four variants 
in Table~\ref{tab:pckvlog}. On cropped evaluations that match the
actual distribution of consumer video, our approach produces substantial
improvement, increasing performance overall for both HMR and CMR.
On the {\it uncropped images} where the head
of the person is visible (which is closer to distributions seen on e.g., 
MPII), our approach remains approximately the same for HMR and actually improves by 
8.6\% for CMR. We note our method underperforms within cropped images on HMR.
There are two reasons for this: first, supervising on out-of-image keypoints encourages
predictions outside of images, sacrificing marginal in-image performance gains. Second, the cost of
supervising on self-generated keypoints is reduced precision in familiar settings.
Nevertheless, CMR improves enough using semi-supervision to still increase on in-image-cropped keypoints.

\subsection{Generalization Evaluations}
\label{subsec:generalization}

\begin{figure*}[t!]
	\centering
			{\includegraphics[width=\textwidth]{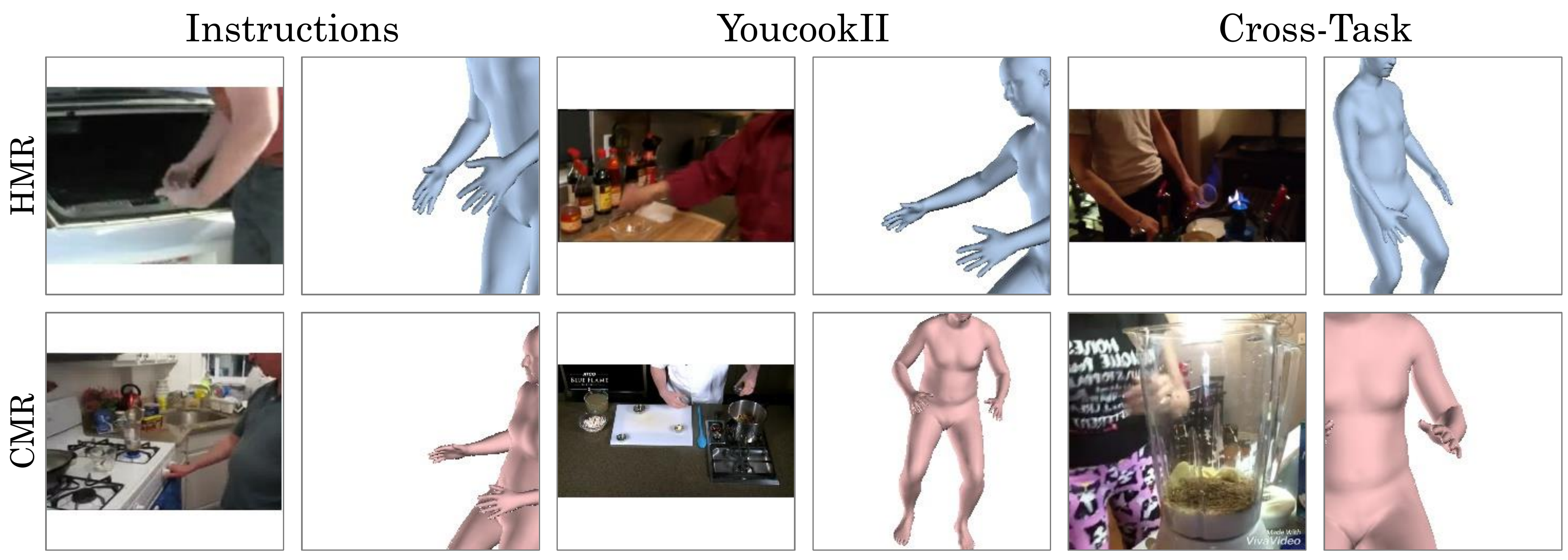}}
	\captionof{figure}{{\bf Results on External Datasets}. While our method trains on Internet Vlogs, performance generalizes to other
	Internet video consisting of a variety of activities and styles; specifically instructional videos and cooking videos.
    }
	\label{fig:fig9}
\end{figure*}

We now test generalization to other datasets. Specifically, we take the
approaches evaluated in the previous section and apply them directly to
Instructions \cite{Alayrac2016}, YouCookII \cite{ZhXuCoCVPR18}, and
Cross-Task \cite{Zhukov19} {\it with no further training}. This tests whether
the additional learning is simply overfitting to VLOG.
We show qualitative results of our system applied to these datasets in
Fig. ~\ref{fig:fig9}. Although the base models also work poorly on these
consumer videos, simply training on VLOG is sufficient to produce 
more reasonable outputs.

\begin{table}[t]
    \caption{{\bf Percentage of Good Meshes on External Datasets}, as judged by human workers. We report results on {\it All} images and in the case of few visible keypoints.}
    \label{tab:good_other}
\resizebox{\ifdim\width>\columnwidth
        \columnwidth
      \else
        \width
      \fi}{!}{
\begin{tabular}{l @{~~~~} cc @{~~~~} cc @{~~~~} cc @{~~~~} cc @{~~~~} cc @{~~~~} cc} 
\toprule
        & \multicolumn{4}{c}{Instructions \cite{Alayrac2016}} 
        & \multicolumn{4}{c}{YouCook II \cite{ZhXuCoCVPR18}} 
        & \multicolumn{4}{c}{Cross-Task \cite{Zhukov19}} 
\\
        & \multicolumn{2}{c}{HMR} & \multicolumn{2}{c}{CMR}
        & \multicolumn{2}{c}{HMR} & \multicolumn{2}{c}{CMR}
        & \multicolumn{2}{c}{HMR} & \multicolumn{2}{c}{CMR}
\\
        & 1-6 & All & 1-6 & All & 1-6 & All & 1-6 & All 
        & 1-6 & All & 1-6 & All 
        \\
\midrule

    Base    & 10.7 & 42.2 & 7.4 & 30.9 
            & 15.9 & 54.6 & 8.5 & 41.8 
            & 13.6 & 52.6 & 7.7 & 37.9 
\\
    Crops   & 25.5 & 53.8 & 28.8 & 52.5 
            & 24.9 & 60.8 & 24.3 & 60.2 
            & 22.3 & 59.0 & 21.5 & 57.7 
\\
    Full    & \bf 37.3 & \bf 60.5 & \bf 35.2 & \bf 57.0 
            & \bf 43.4 & \bf 71.5 & \bf 39.9 & \bf 68.5
            & \bf 37.0 & \bf 68.1 & \bf 31.0 & \bf 63.5 
\\ \bottomrule
\end{tabular} 
}
\end{table}

\noindent {\bf 3D Mesh Judgments:}
This is substantiated quantitatively across the full dataset since, as shown in Table
\ref{tab:good_other}, HMR and CMR perform poorly out-of-the-box.
Our approach, however, can systematically improve their performance without any
additional pose annotations: gains over the best baseline range from 4.5
percentage points (CMR tested on Instructions) to 10.7 percentage points (HMR
tested on YouCookII). Our outputs are systematically preferred by
humans in A/B tests (Table \ref{tab:new_ab}): our approach is $4.6$x -- $8.9$x more likely to be picked
as preferable compared to the base system than the reverse,
and similarly $2.4$x -- $7.8$x more likely to be picked as preferable to
crops than the reverse. 

\noindent {\bf 2D Keypoints:}
Finally, we evaluate PCK. Our approach produces strong performance gains
on two out of the three datasets (YouCookII and Cross-Task), while its performance
is more mixed on Instructions {\it relative to MPII Crops}. We hypothesize 
the relatively impressive performance of MPII Crops is due 
to 40\% of this dataset consisting of car mechanical fixes. These 
videos frequently feature people bending down, for instance 
while replacing car tires. Similar activities such as swimming are more common in MPII than VLOG.
The corresponding array of outdoor scenes also provides less context to accurately infer out-of-image body parts.
Yet, strong human judgment results (Table \ref{tab:good_other}, \ref{tab:new_ab}) indicate training on VLOG
improves coarse prediction quality, even in this setting.

\subsection{Additional Comparisons}
\label{subsec:confidence}

To validate our choice of confidence, we consider two alternative criteria for selecting confident images: agreement between HMR and CMR SMPL parameters, and agreement
between HMR and Openpose \cite{cao2018openpose} keypoints. For fair comparison, implementations closely match our confidence method;
full details and tables are in supplemental. 
Compared to both, our system does about the same or better across datasets, but does not require running two systems.
Agreement with CMR yields cropped keypoint accuracy of 1.5-2.7\% lower, and uncropped accuracy of 0.6\% higher - 0.6\% lower.
Agreement with Openpose is stronger on uncropped images: 0.3\%-2.4\% higher, but weaker on uncropped: 1.3\%-3.5\% lower.

\begin{table}
    \caption{{\bf A/B testing on All Datasets}, using HMR. For each entry we report how frequently (\%) the row wins/ties/loses to the
column. For example, row 2, column 6 shows that our full method is preferred 47\% of the time over a method trained on MPII Crops,
and MPII Crops is preferred over the full method just 6\% of the time}
\label{tab:new_ab}
\resizebox{\ifdim\width>\columnwidth \columnwidth \else \width \fi}{!}{
\begin{tabular}{l@{~~}c@{~~}c@{~~~}c@{~~}c@{~~~}c@{~~}c@{~~~}c@{~~}c} \toprule
Method & \multicolumn{2}{c}{VLOG \cite{Fouhey18}} & \multicolumn{2}{c}{Instructions \cite{Alayrac2016}} & \multicolumn{2}{c}{YouCookII \cite{ZhXuCoCVPR18}} & \multicolumn{2}{c}{Cross-Task \cite{Zhukov19}} 
\\
    & Base  & Crops & Base      & Crops & Base & Crops & Base & Crops 
\\
\midrule
Crops   & 53/28/19 & - & 56/28/16    & -     & 49/35/16 & - & 46/39/15 
\\
Full    & 63/23/15 & 45/43/12 & 65/21/14    & 40/43/17 & 62/32/7 & 47/47/6 & 57/36/7 & 41/53/7
\\
\bottomrule
\end{tabular}
}
\end{table}

\begin{table}[t]
\caption{{\bf PCK @ 0.5 on External Datasets}. We compute PCK in test set images 
in which the head is fully visible. 
These images are then cropped to emulate the keypoint
visibility statistics of the entire dataset, on which we can
calculate PCK on predictions outside the image.
}
\setlength{\tabcolsep}{5pt} 
\centering 
\resizebox{\ifdim\width>\columnwidth
        \columnwidth
      \else
        \width
      \fi}{!}{
  \begin{tabular}{ l c c c c c c c c c c c c c c c c c c c c c c c c } \toprule

Method  & \multicolumn{4}{c}{Instructions \cite{Alayrac2016}}
        & \multicolumn{4}{c}{YouCookII \cite{ZhXuCoCVPR18}}
        & \multicolumn{4}{c}{Cross-Task \cite{Zhukov19}}
\\
        & \multicolumn{2}{c}{HMR \cite{KanazawaHMR17}} 
        & \multicolumn{2}{c}{CMR \cite{Kolotouros19}}
        & \multicolumn{2}{c}{HMR \cite{KanazawaHMR17}} 
        & \multicolumn{2}{c}{CMR \cite{Kolotouros19}}
        & \multicolumn{2}{c}{HMR \cite{KanazawaHMR17}} 
        & \multicolumn{2}{c}{CMR \cite{Kolotouros19}}
        \\
        & Total   & Out 
        & Total   & Out 
        & Total   & Out 
        & Total   & Out 
        & Total   & Out 
        & Total   & Out 

        \\ \midrule
Base    & 42.0      & 19.6  & 32.8  & 17.1
        & 56.0      & 27.7  & 44.0  & 26.9
        & 56.1      & 20.3  & 44.1  & 19.8
\\ 
MPII Crops 
        & \bf 50.6  & 33.7  & \bf 47.9  & \bf 33.9
        & 65.8      & 45.0  & 65.0  & 48.6
        & 62.9      & 32.5  & 61.9  & 38.2
\\
Ours         
        & 48.7      & \bf 36.4  & 44.8  & 33.7
        & \bf 76.7  & \bf 64.1  & \bf 70.7  & \bf 58.5
        & \bf 74.5  & \bf 57.2  & \bf 66.9  & \bf 47.9
\\ \bottomrule
\end{tabular}
\vspace{-0.1in}
}
\label{tab:pck_combined}
\end{table}

We additionally consider performance of our model to the model after only the first iteration of VLOG training,
through A/B testing (full table in supplemental). In all four datasets, the final method is $1.7$x -- $2.8$x more likely to be picked as preferable
to the model after only one round than the reverse.

\section{Discussion} \label{sec:discussion}
We presented a simple but effective approach for adapting
3D mesh recovery models to the challenging world of Internet videos. 
In the process, we showed that current methods 
appear to work poorly on Internet videos, presenting a new opportunity.
Interestingly, while CMR outperforms HMR on Human3.6M, the opposite is true on this new data,
suggesting that performance
gains on standard pose estimation datasets do not always translate into
performance gains on Internet videos. Thanks to the new annotations across the
four video datasets, however, we can quantify this. These keypoint metrics are
validated as a measure for prediction quality given general agreement with
human judgement metrics in extensive testing. We see getting systems to work
on consumer videos, including both the visible and out-of-image
parts, as an interesting and impactful challenge and believe
our simple method provides a strong baseline for work in this
area.

\noindent {\bf Acknowledgments:} 
This work was supported by the DARPA Machine Common Sense Program. We thank
Dimitri Zhukov, Jean-Baptiste Alayrac, and Luowei Zhou, for allowing
sharing of frames from their datasets, and Angjoo Kanazawa and
Nikos Kolotouros for polished and easily extended code.
Thanks to the members of Fouhey AI Lab and Karan Desai for the
great suggestions!

\bibliographystyle{splncs04}
\bibliography{local}

\appendix
\newpage
\section{Method}

\subsection{HMR Network Architecture}

Our HMR network architecture is the same as the HMR model of Kanazawa \textit{et al.} It consists of a ResNet-50
feature extractor, followed by a 3D regression refinement network, consisting of 3 fully-connected 
layers mapping to SMPL, global rotation, and camera parameters $\Theta$. The fully-connected layers concatenate image features, mean SMPL parameters $\ThetaB$, and default global rotation $\RB$ and camera parameters [$\tB, \sB$]. FC layers also use residual links and dropout. More details can be found in the original paper. Also like HMR, we use same-padding for image inputs, although for illustrative purposes images in the paper are shown with white or black padding. 

\subsection{CMR Network Architecture}

Our CMR network architecture is the same as the CMR model of Kolotouros \textit{et al.} It consists first of a ResNet-50 encoder, with the final fully-connected layer removed. This outputs a 2048-D feature vector, which is attached to 3D coordinates of template mesh vertices. A series of graph convolutions then map to a single 3D mesh vertex set, and to camera parameters [$\tB, \sB$]. Finally, a multi-layer perceptron maps these vertices to SMPL parameters $\ThetaB$ and global rotations $\RB$. Final predictions use camera parameters from graph convolutions [$\tB, \sB$], and SMPL parameters and global rotations [$\ThetaB, \RB$] from the MLP. More details can be found in the original paper.

\subsection{Confident Predictions}

As detailed in the paper, predictions are considered confident if average variance of joint rotation parameters across jittered images is less than 0.005 for HMR, chosen empirically. For simplicity, threshold for CMR is chosen so that approximately the same number of confident images are chosen, resulting in a threshold of 0.004. The five images from which predictions are averaged are:
\begin{enumerate}
	\item the original image
	\item the image translated 10 pixels to the top left, padded on the bottom right
	\item the image translated 20 pixels to the top left, padded on the bottom right
	\item the image translated 10 pixels to the bottom right, padded on the top left
	\item the image translated 20 pixels to the bottom right, padded on the top left
\end{enumerate}
\subsection{Cropping}

\begin{figure*}[t!]
	\centering
			{\includegraphics[width=\textwidth]{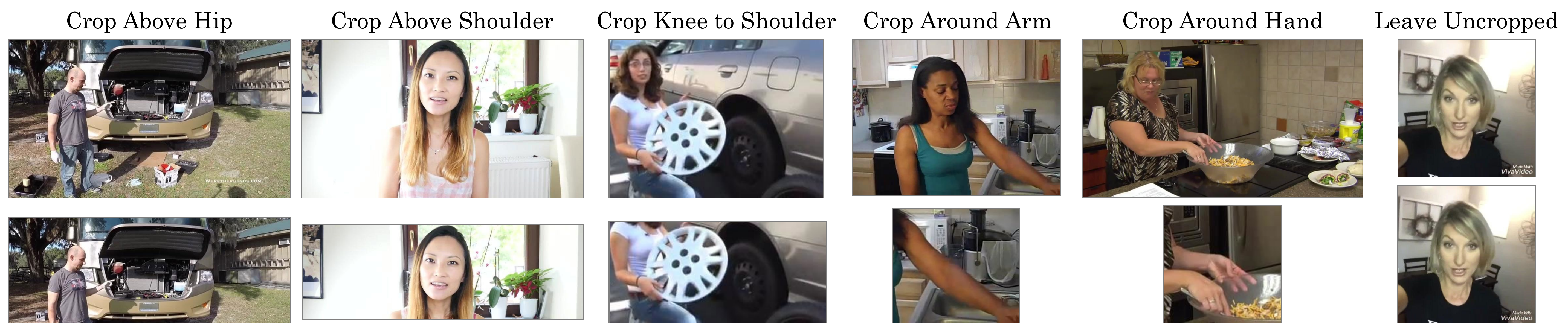}}
	\captionof{figure}{{\bf Generating Cropped Sets.} Training and cropped testing use keypoints to crop images to target visibility specifications. Examples of each crop specification we use are pictured. Some images are left uncropped, and sometimes predefined crops do not further crop images (right).}
\label{fig:crops}
\end{figure*}

During training, the proposed method crops training and validation images into 
five categories: above hip, above shoulders, from knee to shoulder, around only an arm, and around only a hand. 
We show examples of cropping in Fig. \ref{fig:crops}. 
These crops correspond to common crops occurring in consumer video, displayed in Fig. 2 of the paper. 
Above hip corresponds to 
``Legs Not Visible'', knee to shoulder corresponds to ``Head Not Visible'', and above shoulders corresponds
to ``Torso Not Visible''. For brevity, in Fig. 2, we condensed only an arm or only a hand into one image: ``Arms / Hands Visible''.

During training, we sample crops with approximately the same frequency as they occur in
the VLOG validation set.
Proportions are: above
hip in 29\% of images, knee to shoulder in 10\% of images, above shoulders in 16\% of images,
around one arm only in 9\% of images, around one hand in 13\% of images, and 23\% of images
we leave uncropped. 
On both MPII and VLOG, for both models, we crop to our target crops using keypoints. 
Ground truth keypoints are used on MPII, 
and reprojected keypoints from confident models are used on VLOG.
Above hip crops are made from the lower of the hip keypoints. Knee to shoulder crops use
the higher of knee keypoints to the bottom of shoulder keypoints. Above
shoulder crops use the lower of shoulder and neck as the bottom of the crop. Elbow and 
wrist keypoints are used to approximate one arm and one hand crops. 
If keypoints used for cropping are outside of images, for simplicity we presume the 
image is already cropped and do not crop further. If a prospective crop would be smaller than 30 pixels, we also do not crop to prevent training on very low-resolution examples.

\section{Datasets}

\subsection{Dataset Annotation}

As detailed in the paper, for each of VLOG, Instructions, YoucookII, and Cross-Task; we subsample 
5k random frames containing exactly one person. Next, we use human annotators to label human keypoints on all of these frames. The full test sets consist of images in which at least one keypoint is annotated, on which workers can agree (details in Keypoint Annotation). They are of size 4.1k, 2.2k, 3.3k, and 3.9k, for VLOG, Instructions, YoucookII, and Cross-Task, correspondingly. 
Instructions is notably the smallest because it has a higher proportion of images containing no human keypoints.
The most common instance of this occurring is when some of a hand is visible, which does not contain a keypoint, but a wrist is not visible, which corresponds to a keypoint.

All human annotations were gathered using thehive.ai, a website similar to Amazon Mechanical Turk. 
Annotators were given instructions and examples of each category. Then, they were given a test to identify whether they understood the task. Workers that passed the test were allowed to annotate images. However, as they annotated, they were shown, at random, gold standard sentinel examples (i.e., that we labeled), that tested their accuracy. The platform automates the entire process.
Because some workers spoke only Spanish, we put directions in both English and Spanish. English annotation instructions are provided in Subsections \ref{sec:kp}, \ref{sec:mesh}, and \ref{sec:ab}.

\noindent{\bf Keypoint Annotation} 

Although traditionally disagreement on labeling ground truth is handled by thehive.ai, the company does
not currently support labeling keypoints in the instance there are an ambiguous number
of keypoints visible, which occurs here. For instance, consider the case where a person's elbow
is at the edge of an image. Some annotators may label the elbow while others do not.
To deal with annotations containing different number of visible joints, we combine 
predictions between workers on a given image ourselves, using the
median number of joints annotated between workers, and average the locations.

Specifically, we have each image labeled three separate times. If two or three of the occurrences of the image
see no joints labeled, we consider it as having no joints visible. In other cases, we take the median
number of joints visible between the three instances, and average these joints across instances when
possible. If any given joint has predictions differing by a large margin (10\% of image size), we do not
annotate it. If all three workers disagree on number of joints visible, we consider it ambiguous and do not add it to our test set.

\noindent{\bf 3D Mesh Annotation} 

\noindent{\it Absolute Judgment:} Workers annotated whether the mesh was largely correct or not. The percent of times they annotated correct gives Percentage of Good Meshes.
For mesh scoring, all ablations were scored at the same time, to avoid possible
bias between scores between models. Additionally, model outputs and images were put in random order 
to avoid one person seeing many outputs from the same model in a row.

\noindent{\it Relative Judgment:} For A/B testing, workers were presented with an image and two outputs. They selected the output that best matched the image.
Order in which predictions were seen 
in relation to the image was randomized, and which outputs were compared was 
also presented in random order.

\subsection{Dataset Details and Statistics}

As explained in the paper, we evaluate keypoint accuracy on images in which the head is visible in order to calculate PCK. This results in test subsets of size 1.8k, 0.8k, 1.5k, and 1.9k, for VLOG, Instructions, YoucookII, and Cross-Task, correspondingly. These sets are not representative of the full test-set visibility statistics, and do not allow for out-of-image keypoint evaluation. Therefore, we use cropping of body-parts to closely match aggregate test set statistics. We use the same canonical crops as during training, displayed in Fig. \ref{fig:crops}. However, we explicitly choose crop proportions to closely match full test sets. 

Uncropped keypoint test sets are biased since their images always contain head keypoints; needed for computing PCK. Therefore, we must crop aggressively to match full test set statistics. 
Furthermore, above hip and above shoulder crops are not useful to this end, as they include head keypoints. Knee to shoulder keypoints also are not optimal as they exclude leg keypoints too often,
while continuing to sometimes include shoulder and neck keypoints. Instead, to match full test set statistics, we utilize crops around hands and arms frequently, while leaving some images uncropped.
Statistics on full test sets, uncropped keypoint test sets, and cropped keypoint test sets are detailed in Table \ref{tab:tab1}.

\begin{table}[]
\caption{{\bf Proportion of Visible Joints in Test Sets}. Proportion of dataset images containing 
a particular joint for each of: Uncropped Keypoint Test Set (Uncr.), Cropped Keypoint Test Set (CR.), and Full Test Set (Full). Also, mean number of keypoints (Keypoints) per image.}
\begin{tabular}{l @{~~~} ccc @{~~~~~} ccc @{~~~~~} ccc @{~~~~~} ccc}
\toprule
         & \multicolumn{3}{c}{VLOG} & \multicolumn{3}{c}{Instructions} & \multicolumn{3}{c}{YoucookII} & \multicolumn{3}{c}{Cross-Task} \\
         & Uncr.    & Cr.    & Full   & Uncr.       & Cr.      & Full      & Uncr.      & Cr.     & Full     & Uncr.      & Cr.      & Full     \\
          \midrule
R Ank    & 0.16  &  0.07  & 0.11   & 0.22            & 0.11         & 0.13      & 0.00           & 0.00        & 0.00     & 0.04           & 0.02         & 0.03     \\
R Kne    & 0.29  &  0.16  & 0.20   & 0.32            & 0.23         & 0.24      & 0.01           & 0.00        & 0.01     & 0.07           & 0.04         & 0.06     \\
R Hip    & 0.48  & 0.30  & 0.34    & 0.64            & 0.40         & 0.32      & 0.51           & 0.33        & 0.30     & 0.46           & 0.30         & 0.31     \\
L Hip    & 0.48   & 0.31  & 0.34    & 0.66            & 0.43         & 0.33      & 0.51           & 0.35        & 0.30     & 0.46           & 0.30         & 0.31     \\
L Kne    & 0.29   & 0.16  & 0.20   & 0.33            & 0.23         & 0.25      & 0.01           & 0.00        & 0.01     & 0.07           & 0.04         & 0.06     \\
L Ank    & 0.15   & 0.07 & 0.10    & 0.23            & 0.11         & 0.15      & 0.00           & 0.00        & 0.00     & 0.04           & 0.02         & 0.04     \\
R Wri    & 0.78   &  0.65   & 0.73    & 0.83            & 0.69         & 0.76      & 0.78           & 0.61        & 0.72     & 0.76           & 0.59         & 0.72     \\
R Elb    & 0.75    & 0.48    & 0.55    & 0.87            & 0.57         & 0.52      & 0.79           & 0.50        & 0.45     & 0.78           & 0.50         & 0.51     \\
R Sho    & 0.95  &  0.63     & 0.61    & 0.95            & 0.43         & 0.45      & 0.99           & 0.60        & 0.53     & 0.97           & 0.60         & 0.56     \\
L Sho    & 0.95  &  0.62     & 0.61     & 0.94            & 0.43         & 0.44      & 0.99           & 0.58        & 0.54     & 0.97           & 0.60         & 0.56     \\
L Elb    & 0.76    &  0.49   & 0.54     & 0.88            & 0.56         & 0.52      & 0.78           & 0.51        & 0.46     & 0.78           & 0.51         & 0.51     \\
L Wri    & 0.76   &  0.66    & 0.70      & 0.83            & 0.70         & 0.71      & 0.76           & 0.61        & 0.70     & 0.75           & 0.61         & 0.71     \\
Neck     & 1.00   & 0.67     & 0.61      & 1.00            & 0.46         & 0.45      & 1.00           & 0.62        & 0.54     & 1.00           & 0.63         & 0.57     \\
Head Top & 1.00  &  0.55     & 0.47     & 1.00            & 0.28         & 0.38      & 1.00           & 0.36        & 0.48     & 1.00           & 0.47         & 0.51     \\
Nose     & 0.95   &   0.61   & 0.57     & 0.98            & 0.42         & 0.47      & 1.00           & 0.52        & 0.54     & 0.99           & 0.56         & 0.57     \\
L Eye    & 0.95   &  0.58    & 0.55   & 0.98            & 0.38         & 0.45      & 1.00           & 0.47        & 0.54     & 0.99           & 0.54         & 0.56     \\
R Eye    & 0.95  &  0.58     & 0.55     & 0.98            & 0.36         & 0.45      & 1.00           & 0.48        & 0.54     & 0.99           & 0.53         & 0.56     \\
L Ear    & 0.93   &  0.56    & 0.53     & 0.96            & 0.35         & 0.43      & 0.99           & 0.47        & 0.53     & 0.99           & 0.53         & 0.55     \\
R Ear    & 0.93  &  0.56     & 0.53     & 0.94            & 0.33         & 0.42      & 0.99           & 0.47        & 0.53     & 0.98           & 0.53         & 0.55     \\
\midrule
Keypoints & 13.5        & 8.7       & 8.8   & 14.5           & 7.5         & 7.9      & 13.1          & 7.5        & 7.7     & 13.1          & 7.9         & 8.2    \\
\bottomrule
\end{tabular}
\label{tab:tab1}
\end{table}

\newpage
\section{Additional Qualitative Results}

\subsection{Additional Results on VLOG} \label{sec:sec_vlog}

\begin{figure}[htp!]
	\centering
			{\includegraphics[width=\textwidth]{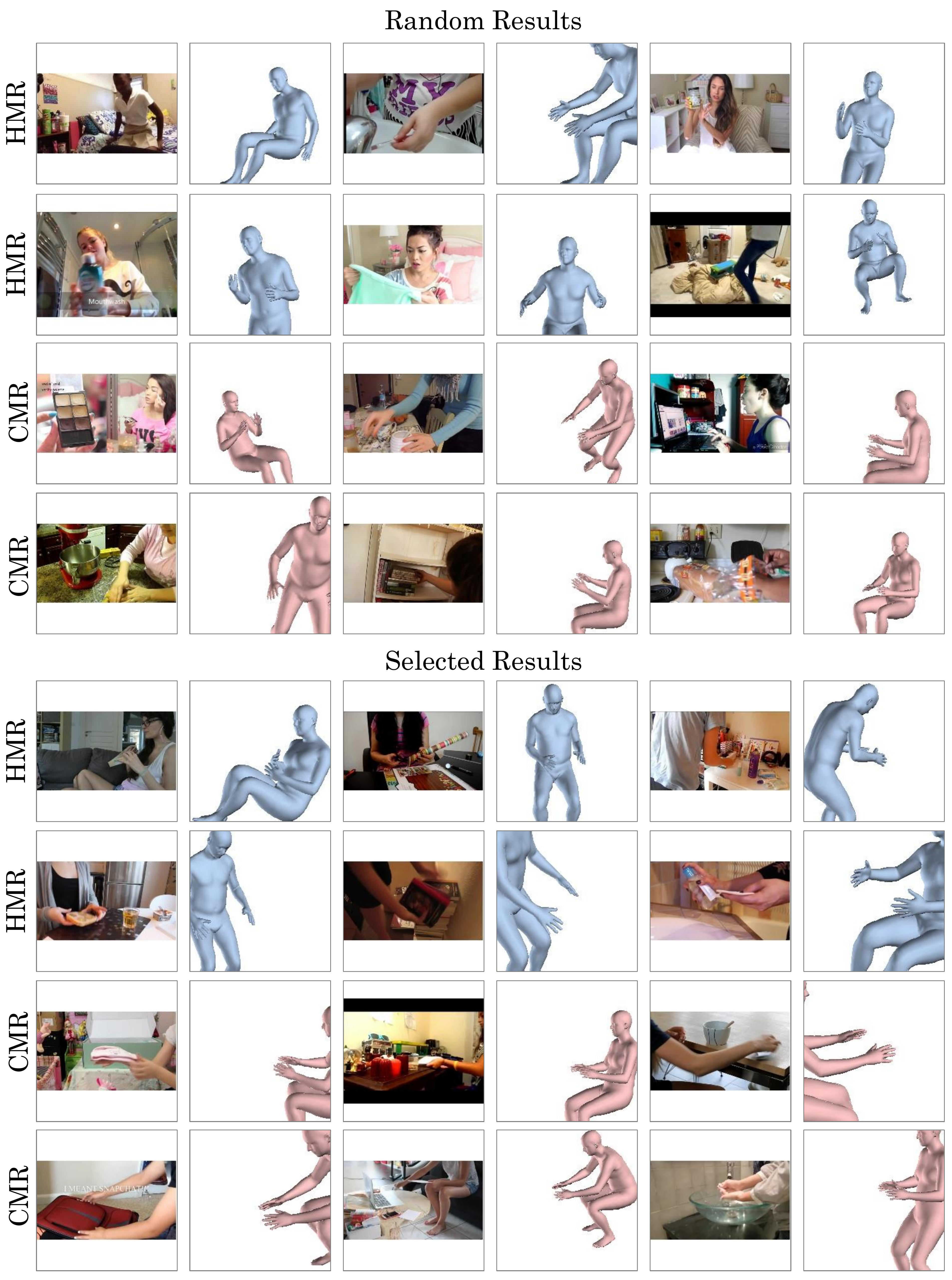}}
\end{figure}

\subsection{Additional Results on Instructions} \label{sec:sec_iv}

\begin{figure}[htp!]
	\centering
			{\includegraphics[width=\textwidth]{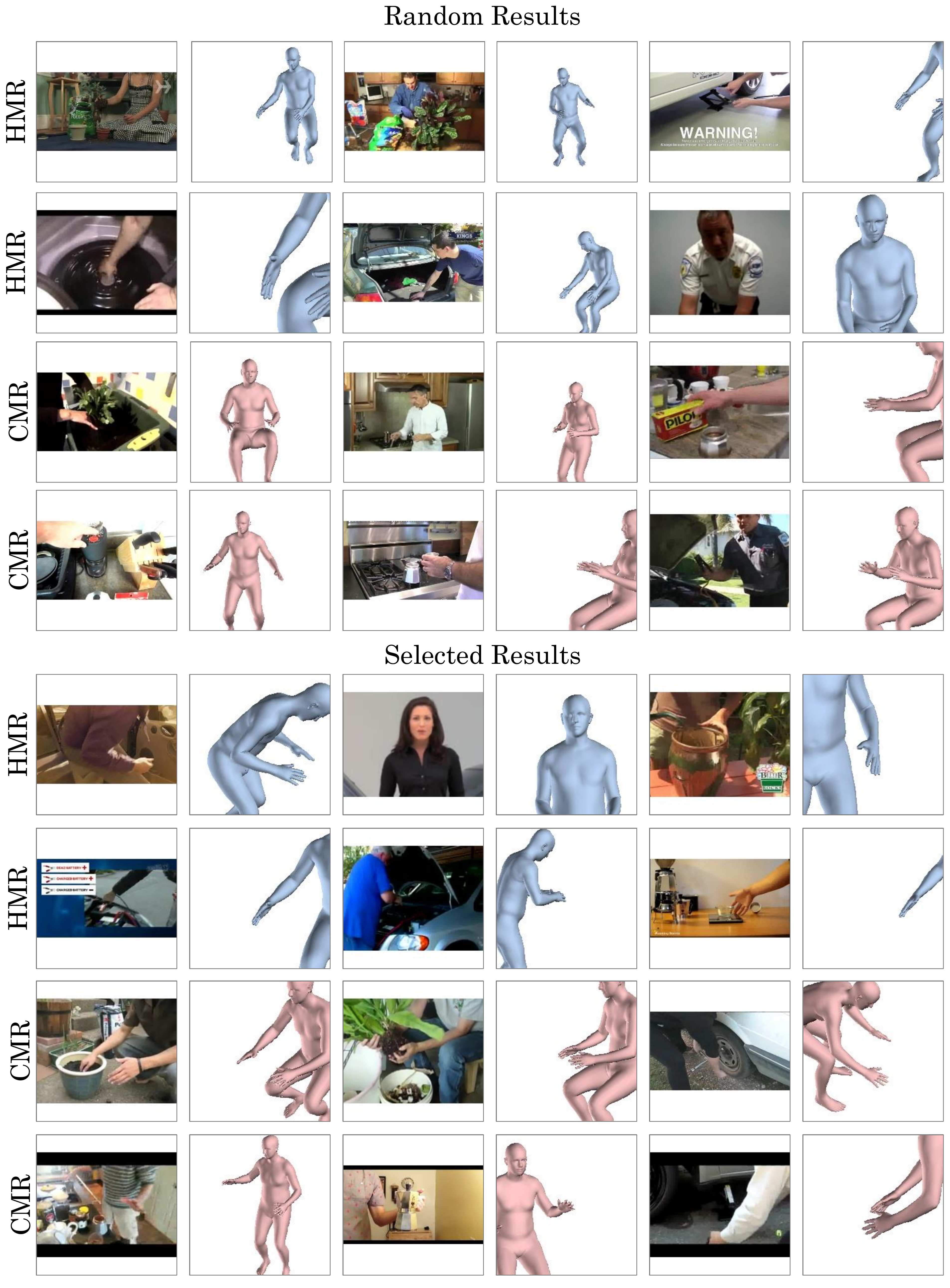}}
\end{figure}

\subsection{Additional Results on YoucookII} \label{sec:sec_yc}

\begin{figure}[htp!]
	\centering
			{\includegraphics[width=\textwidth]{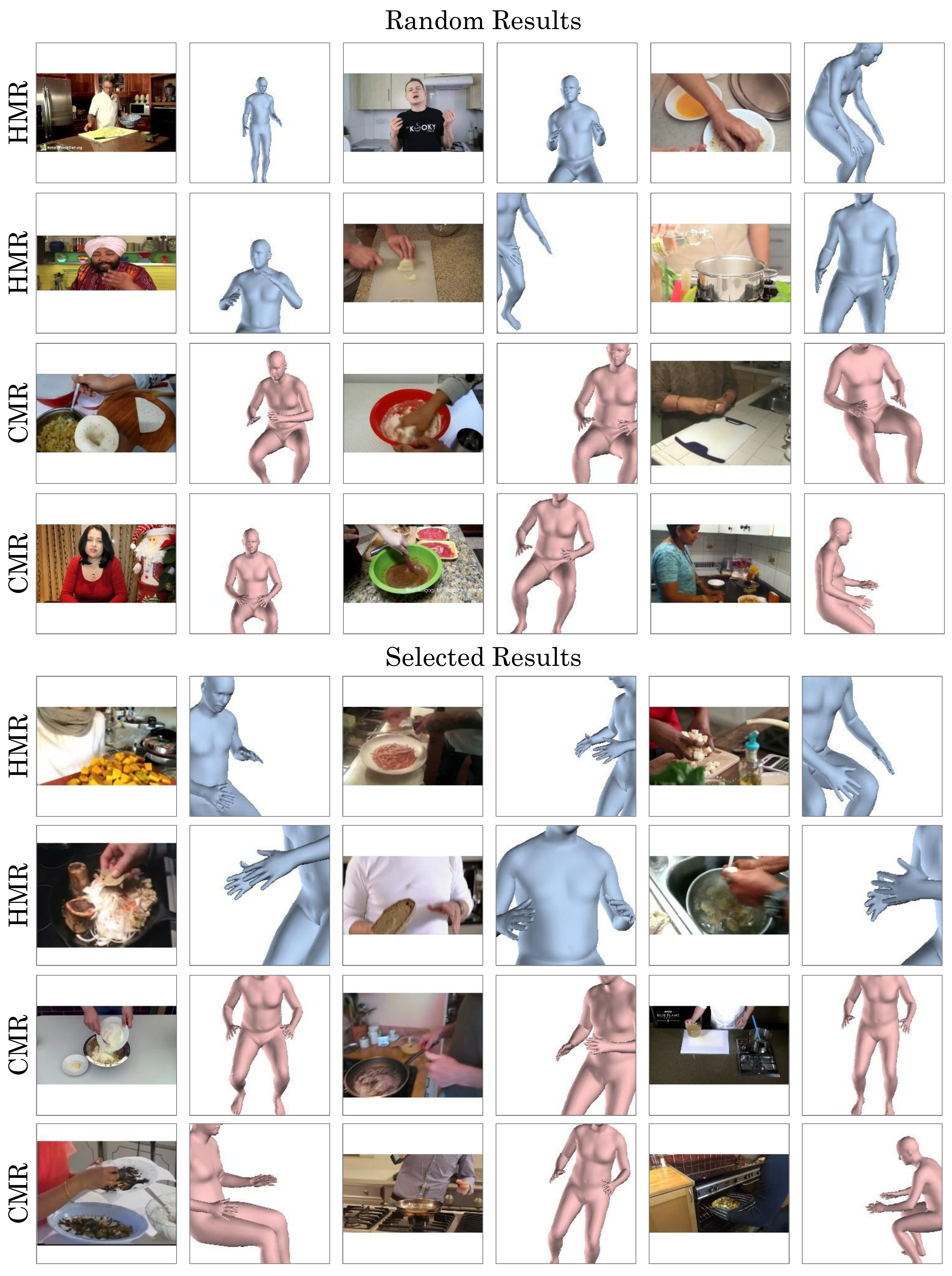}}
\end{figure}

\subsection{Additional Results on Cross-Task} \label{sec:sec_ct}

\begin{figure}[htp!]
	\centering
			{\includegraphics[width=\textwidth]{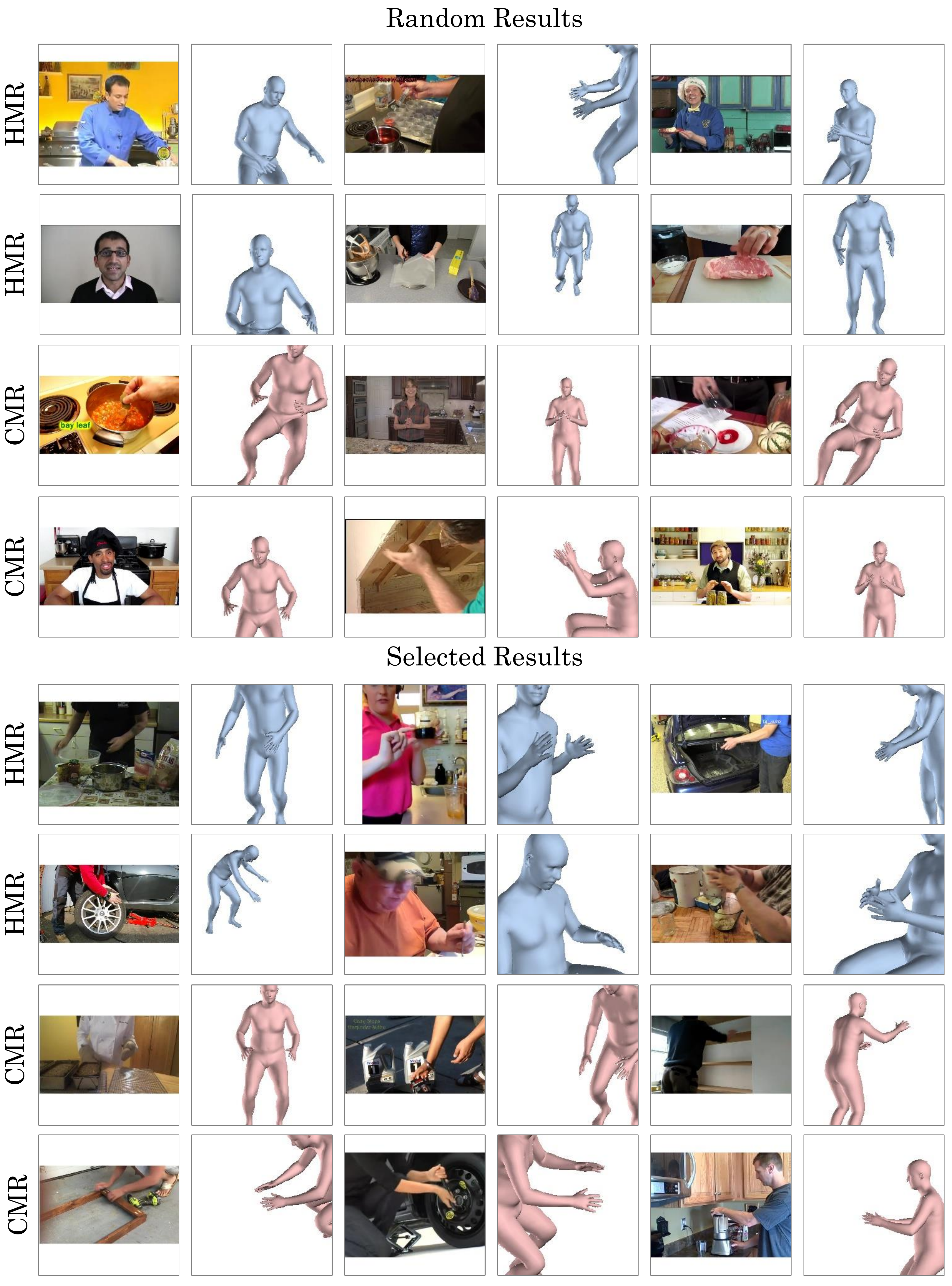}}
\end{figure}
\relax
\section{Additional Results} \label{sec:sec_add}

\noindent {\bf Comparison to One Round:} The paper presents comprehensive comparison between our full method, the method after only MPII crops, and original baselines.
As our method performs two iterations of training on VLOG, we additionally compare final performance to the method after only a single
round of VLOG training in Table \ref{tab:tab_ab}. As reported in the paper, workers prefer the model after a second iteration of training.

\begin{table}[h!]
\centering
    \caption{{\bf A/B testing on All Datasets}, using HMR. For each entry we report how frequently (\%) our full method wins/ties/loses to the method trained with only one round of VLOG training. For example, column 1 shows Full is preferred to One Round 18\% of the time, and One Round is preferred just 8\% of the time.}
\label{tab:tab_ab}
\resizebox{\ifdim\width>\columnwidth \columnwidth \else \width \fi}{!}{
\begin{tabular}{l@{~~}c@{~~}c@{~~~}c@{~~}c@{~~~}c@{~~}c@{~~~}c@{~~}c} \toprule
Method & \multicolumn{4}{c}{One Round}
\\
 & \multicolumn{1}{c}{VLOG} & \multicolumn{1}{c}{Instructions} & \multicolumn{1}{c}{YouCookII} & \multicolumn{1}{c}{Cross-Task} 
\\
\midrule
Full    & 18/74/8 & 15/76/9 & 18/75/7 & 15/80/5
\\
\bottomrule
\end{tabular}
}
\end{table}

\noindent {\bf Comparison to other confidence methods:} As reported in the paper, our selection of confidence also performs similarly to slightly better to agreement between CMR and HMR SMPL joints, and to agreement between HMR keypoints and Openpose keypoints. Full results are available in Table \ref{tab:tab_rebuttal}. CMR / HMR agreement is done in the same manner as our method, and thresholds are set to select approximately the same number of images as our method. Training takes approximately the same time, and two rounds of self-training are used. The same is true of comparison with Openpose, with the distinction Openpose can only predict keypoints inside an image, so we only consider joints both networks predict as in-image. Additionally, Openpose filters out unconfident keypoints, so we only compare joints predicted by both networks. We observe Openpose struggles especially if the face is truncated, so agreement is mostly in highly-visible settings. This leads to better uncropped keypoint accuracy, but worse cropped. 

\begin{table}[h!]
\centering
    \caption{{\bf PCK @ 0.5 on All Datasets}, using HMR. We compute PCK in test set images 
in which the head is fully visible. 
These images are then cropped to emulate the keypoint
visibility statistics of the entire dataset, on which we can
calculate PCK on predictions outside the image.
We also compute PCK on the uncropped images.}
\setlength{\tabcolsep}{5pt} 
\centering 
\resizebox{\ifdim\width>\columnwidth
        \columnwidth
      \else
        \width
      \fi}{!}{
  \begin{tabular}{ l c c c c c c c c c c c c} \toprule

Method  & \multicolumn{3}{c}{VLOG}
	& \multicolumn{3}{c}{Instructions}
        & \multicolumn{3}{c}{YouCookII}
        & \multicolumn{3}{c}{Cross-Task}
        \\
	& \multicolumn{2}{c}{Cropped} & Uncr.     
        & \multicolumn{2}{c}{Cropped} & Uncr.     
	& \multicolumn{2}{c}{Cropped} & Uncr.     
        & \multicolumn{2}{c}{Cropped} & Uncr.     
	 \\
        & Total   & Out & Total 
        & Total   & Out & Total 
        & Total   & Out & Total
        & Total   & Out & Total 
        \\ \midrule
CMR agreement    & 54.3  & 35.9  & 68.1
        & 47.2 & 33.9  & 78.5
        & 74.0 & 59.5  & 94.9
	 & 72.2 & 51.8  & 90.7
\\ 
Openpose agreement  & 54.6 & 34.6 & \bf 71.1
        & 46.1 & 31.8  & \bf 79.8
        & 73.2  & 58.8  & \bf 95.7
	 & 71.3 & 50.0  & \bf 92.2
\\ 
Ours  & \bf 55.9  & \bf 38.9  & 68.7
        & \bf 48.7      & \bf 36.4  & 77.9
        & \bf 76.7  & \bf 64.1  & 95.4 
        & \bf 74.5  & \bf 57.2  & 91.1
\\ \bottomrule
\end{tabular}
\vspace{-0.1in}
}
\label{tab:tab_rebuttal}
\end{table}

\vspace*{0cm}

\section{Annotation Instructions}
\subsection{Keypoint Annotation Instructions} \label{sec:kp}

\begin{tabular}{c}
			{\includegraphics[width=\textwidth,height=.9\textheight,keepaspectratio]{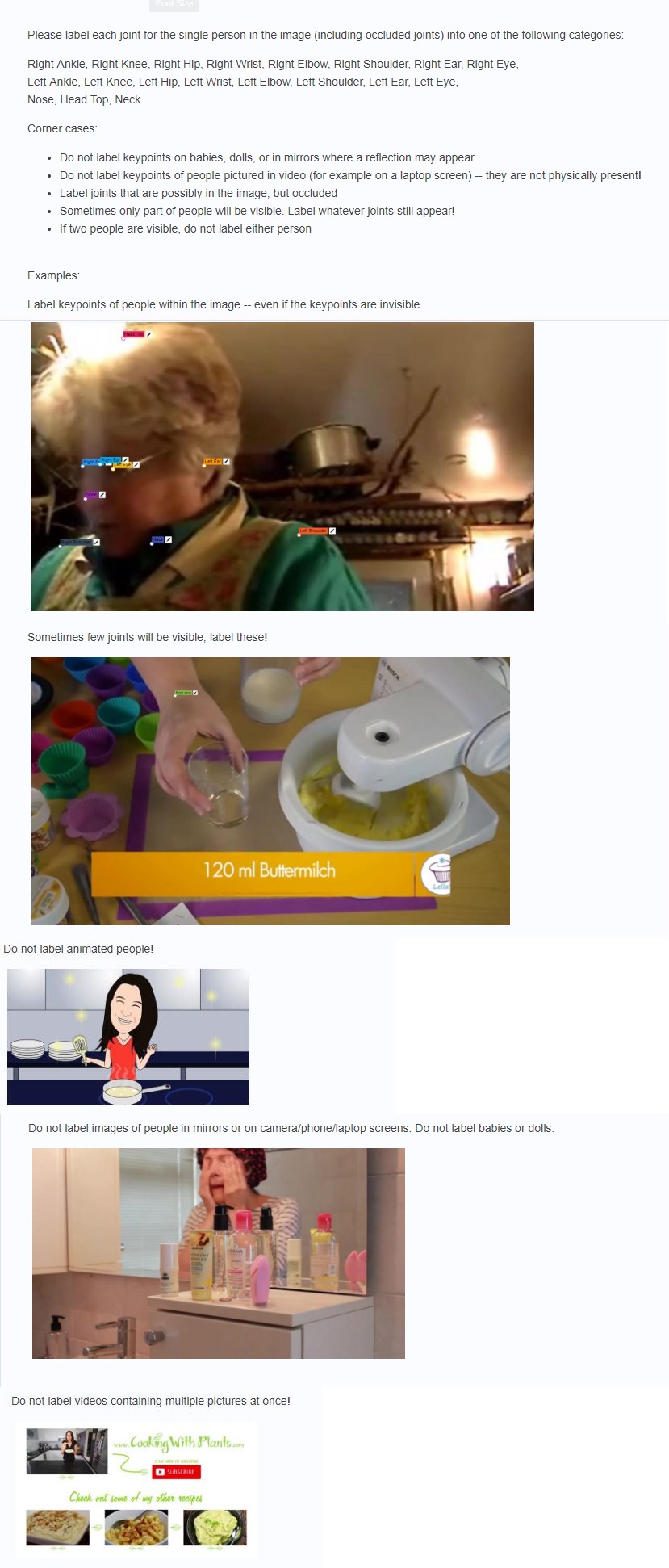}}
\label{fig:kpfig}
\end{tabular}

\clearpage

\subsection{Mesh Scoring Instructions}\label{sec:mesh}

\begin{tabular}{c}
			\includegraphics[width=.5\textwidth, height=\textheight,keepaspectratio]{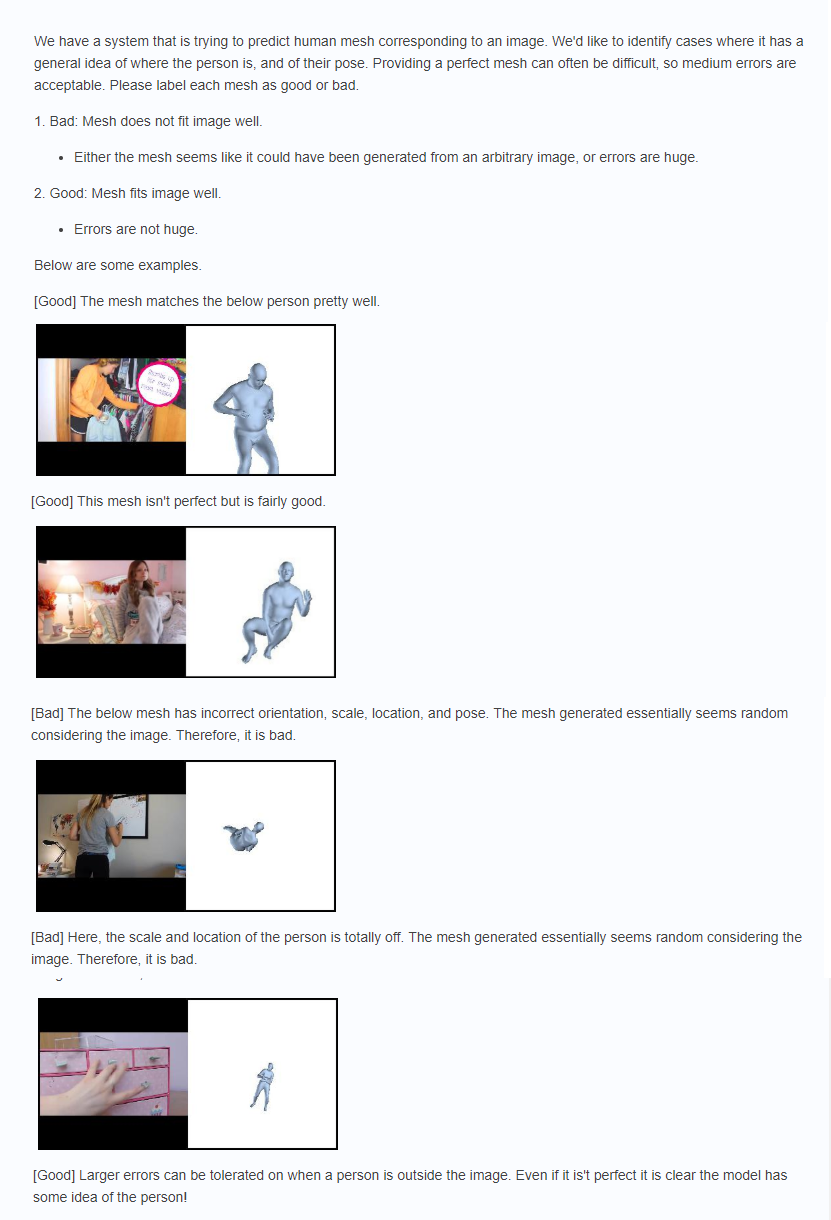} \\ 
			 \includegraphics[width=.5\textwidth, height=\textheight,keepaspectratio]{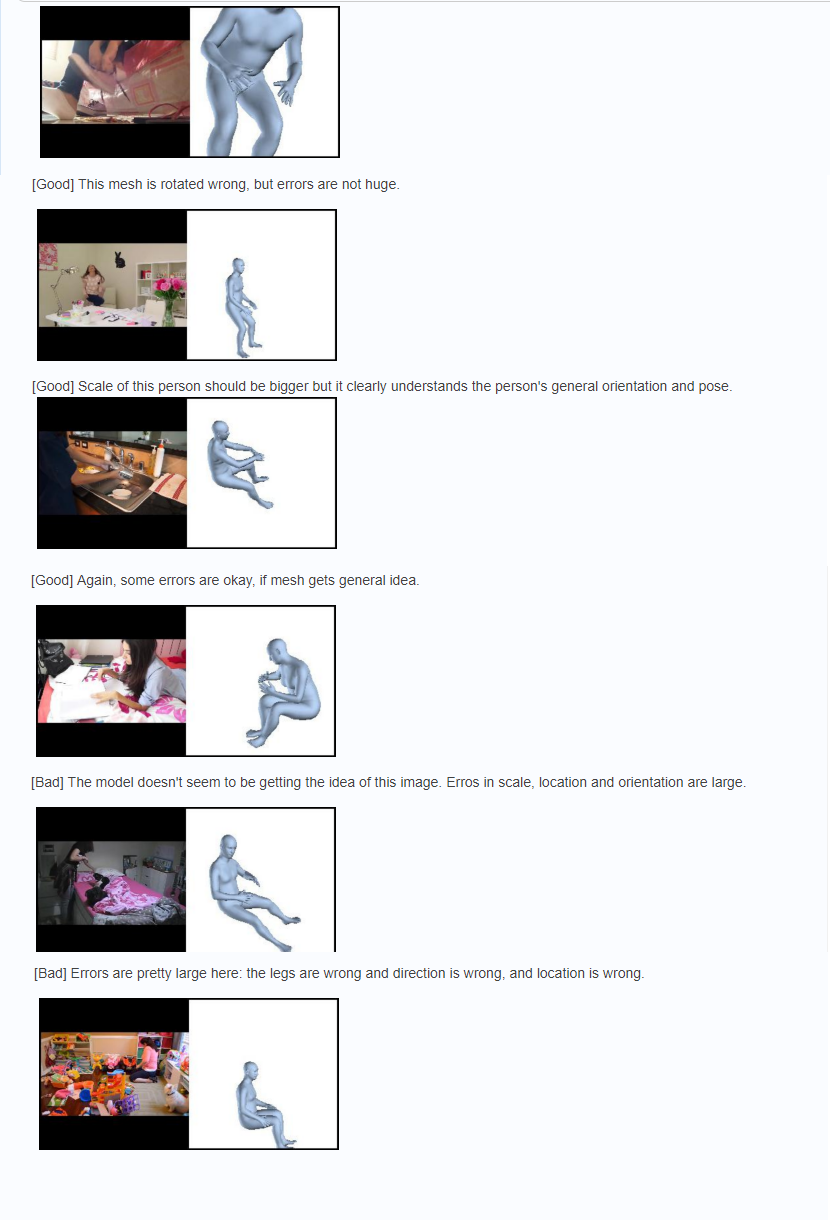} \\
\label{fig:score}
\end{tabular}

\clearpage
\subsection{A/B Testing Instructions}\label{sec:ab}

\begin{tabular}{c}
			{\includegraphics[width=.9\textwidth, height=.9\textheight,keepaspectratio]{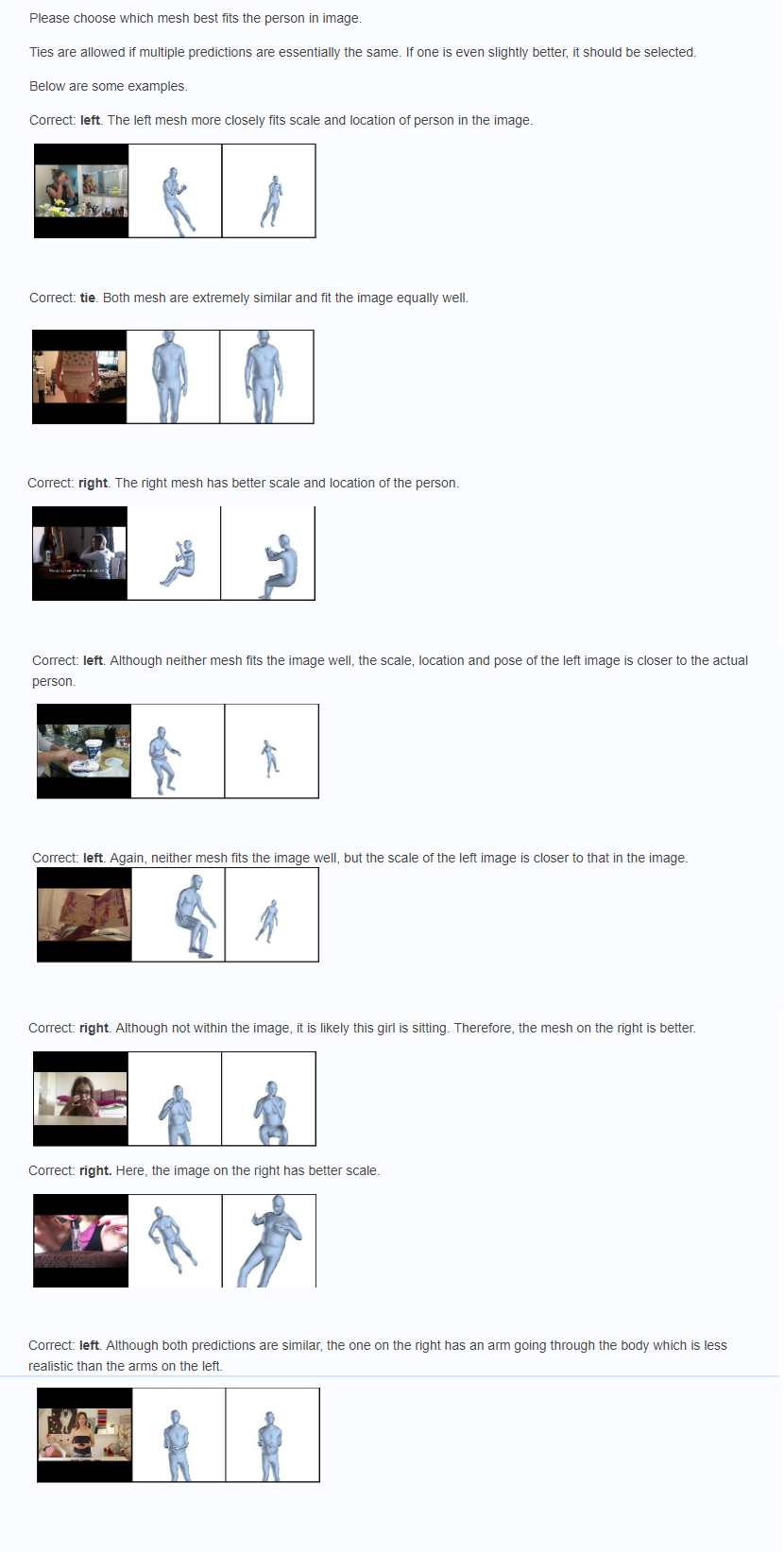}}
\label{fig:ab}
\end{tabular}

\end{document}